%% file: main.tex
\documentclass[10pt,twocolumn,letterpaper]{article}

\usepackage[pagenumbers]{cvpr} % To force page numbers, e.g. for an arXiv version

\usepackage{algorithm}
\usepackage{algpseudocode}
\usepackage{amsmath}
\usepackage{float}
\floatstyle{plaintop}
\restylefloat{table}
\usepackage{enumitem}
\usepackage{comment}
\usepackage{bbding}
\usepackage{placeins}
\usepackage[table]{xcolor}
\usepackage{tabularx}
\usepackage{adjustbox}
\usepackage{xspace}
\usepackage[acronym]{glossaries}
\usepackage{wrapfig}
\usepackage{mwe}
\usepackage{subcaption}
\usepackage{soul}
\usepackage{bm}
\usepackage{amssymb}
\usepackage{microtype}
\usepackage{mathtools}
\usepackage{ragged2e}

\definecolor{cvprblue}{rgb}{0.21,0.49,0.74}
\usepackage[pagebackref,breaklinks,colorlinks,allcolors=cvprblue]{hyperref}
\usepackage{graphicx}
\usepackage{multirow}
\usepackage{booktabs}
\usepackage{array}
\usepackage{gensymb}
\usepackage{physics}
\usepackage{makecell}
\usepackage{pifont}

\input{math_commands.tex}

\definecolor{bblue}{rgb}{0,150,230}
\definecolor{mygray}{gray}{.9}
\definecolor{lightgray}{gray}{.96}
\definecolor{myy}{RGB}{126,95,0}
\definecolor{ggray}{RGB}{127,127,127}
\definecolor{mygreen}{RGB}{93,173,85}
\definecolor{myred}{RGB}{240,16,89}
\definecolor{myblue}{RGB}{0,114,188}
\definecolor{darkgreen}{rgb}{0.0, 0.5, 0.0}
\definecolor{demphcolor}{RGB}{100,100,100}
\definecolor{sh_blue}{rgb}{0,0.60,0.93}
\definecolor{sh_red}{rgb}{0.8627, 0.3098, 0.3176}
\definecolor{lightpink}{rgb}{0.918, 0.761, 0.761}
\definecolor{lightblue}{rgb}{0.671, 0.773, 0.863}
\definecolor{customlightgreen}{rgb}{0.8196, 0.8784, 0.7098}
\definecolor{lightpeach}{rgb}{1.0, 0.882, 0.788}
\definecolor{customblue}{rgb}{0.180, 0.400, 0.522}
\definecolor{lightcyan}{rgb}{0.8196, 0.9725, 0.9804}

\hypersetup{
    colorlinks = true,
    linkcolor  = red,
    urlcolor   = magenta,
    citecolor  = myblue,
}

\newcommand{\name}{\textsc{ID-Crafter}}

\def\paperID{13542} % *** Enter the Paper ID here
\def\confName{CVPR}
\def\confYear{2026}

\title{ID-Crafter: VLM-Grounded Online RL \\ for Compositional Multi-Subject Video Generation}

\author{Panwang Pan$^{*\dagger\ddagger}$,  Jingjing Zhao$^{*}$, Yuchen Lin$^{*}$,  Chenguo Lin$^{}$, Chenxin Li$^{}$,  \\ 
 Hengyu Liu$^{}$, Tingting Shen$^{}$, Yadong Mu$^{\ddagger}$ \\ 
 Xiamen University, East China Normal University, Peking University, CUHK \\ 
 $^*$ Equal contribution\quad $^\dagger$ Project lead\quad $^\ddagger$ Corresponding author \\ 
\url{https://angericky.github.io/ID-Crafter}
}

\begin{document}
\maketitle

% --- Migrated Teaser Figure ---
\begin{figure*}[htbp!]
    \centering
    \includegraphics[width=0.95\linewidth]{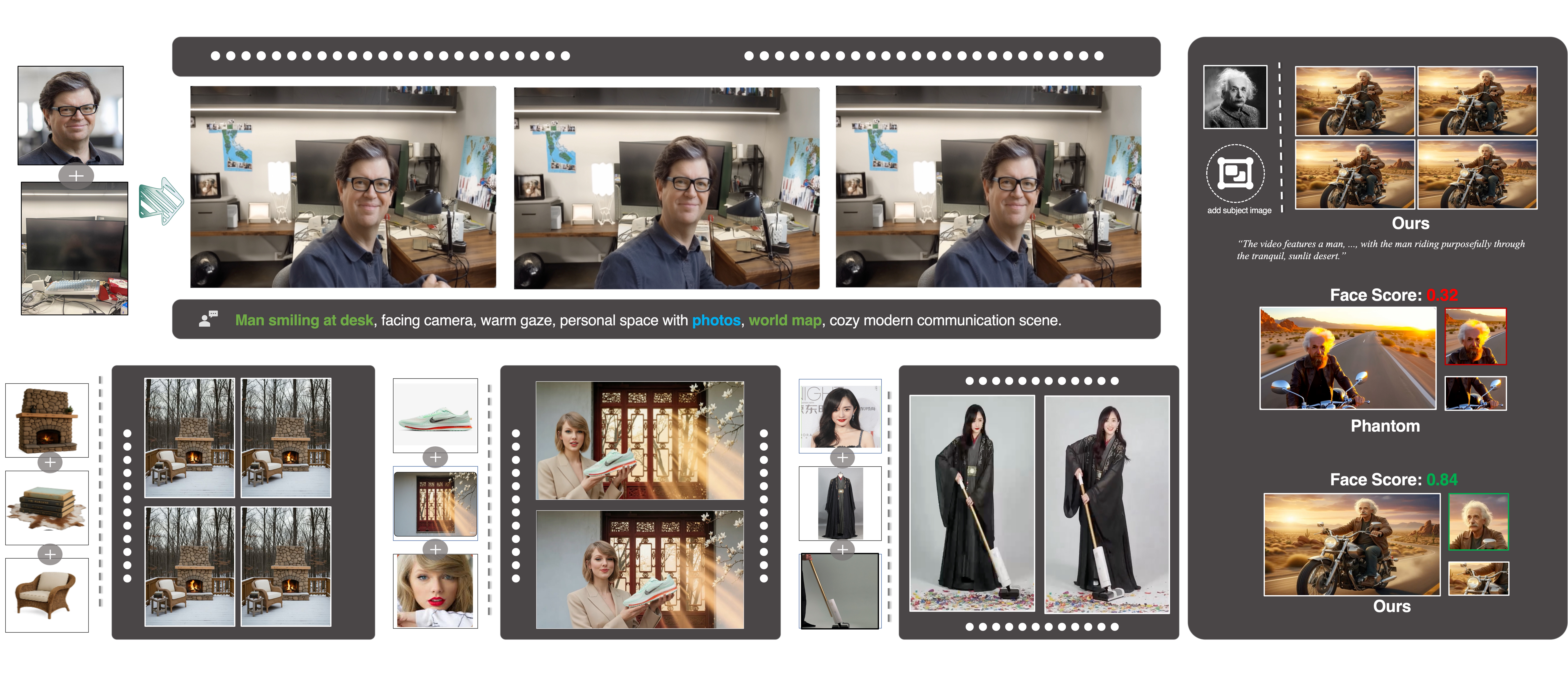}
    \caption{Given a text prompt and multiple reference images, \name{} generates subject-consistent videos and achieves impressive subject ID preservation (e.g., face score) compared with the previous state-of-the-art methods, such as Phantom~\citep{liu2025phantom}.}
    \label{fig:teaser}
\end{figure*}

% --- Migrated Document Body Structure ---
\begin{abstract}
\input{sections/00_abstract}

\end{abstract}

\input{sections/01_introduction}
\input{sections/02_related_work}

\input{sections/03_method}

\input{sections/04_experiments}

\input{sections/05_conclusion}

% --- Migrated Bibliography ---
\clearpage
{
    \small
    \bibliographystyle{ieeenat_fullname}
    \bibliography{references}
}

\clearpage
\appendix
\renewcommand{\thefigure}{S.\arabic{figure}}
\renewcommand{\thetable}{S.\arabic{table}}
\renewcommand{\theequation}{S.\arabic{equation}}
\maketitlesupplementary
\input{sections/99_appendix}

\end{document}

%% file: math_commands.tex
\usepackage{amsmath,amsfonts,bm}

\def\eqref#1{equation~\ref{#1}}
\def\1{\bm{1}}

\def\vv{{\bm{v}}}

\DeclareMathAlphabet{\mathsfit}{\encodingdefault}{\sfdefault}{m}{sl}
\SetMathAlphabet{\mathsfit}{bold}{\encodingdefault}{\sfdefault}{bx}{n}

%% file: sections/00_abstract.tex
% Video generative models pretrained on large-scale datasets can produce high-quality videos, but are often conditioned on text or a single image, limiting controllability and applicability. 
% We introduce \name{}, a novel framework that addresses this gap by tackling multi-subject video generation from a text prompt and reference images. This task is challenging as it requires preserving subject identities, integrating semantics across subjects and modalities, and maintaining temporal consistency.
% To faithfully preserve the subject consistency and textual information in synthesized videos, \name{}~designs a hierarchical identity-preserving attention mechanism, which effectively aggregates features within and across subjects and modalities. 
% To effectively allow for the semantic following of user intention, we introduce 
% semantic understanding via pretrained Vision-Language Model (VLM), leveraging VLM's superior semantic understanding to provide fine-grained guidance and capture complex interactions between multiple subjects. 
% Considering that standard diffusion loss often fails in aligning the critical concepts like subject ID,
% we employ an online reinforcement learning phase to drive the overall training objective of \name{} into RLVR.
% Extensive experiments demonstrate that our model surpasses existing methods in identity preservation, temporal consistency, and video quality. 
% Recent advances in video generative models have achieved high-fidelity in video synthesis.

Significant progress has been achieved in high-fidelity video synthesis, yet current paradigms often fall short in effectively integrating identity information from multiple subjects. This leads to semantic conflicts and suboptimal performance in preserving identities and interactions, limiting controllability and applicability. To tackle this issue, we introduce \name{}, a framework for multi-subject video generation that achieves superior identity preservation and semantic coherence. \name{} integrates three key components: (i) a hierarchical identity-preserving attention mechanism that progressively aggregates features at intra-subject, inter-subject, and cross-modal levels; (ii) a semantic understanding module powered by a pretrained Vision-Language Model (VLM) to provide fine-grained guidance and capture complex inter-subject relationships; and (iii) an online reinforcement learning phase to further refine the model for critical concepts. Furthermore, we construct a new dataset to facilitate robust training and evaluation. Extensive experiments demonstrate that \name{} establishes new state-of-the-art performance on multi-subject video generation benchmarks, excelling in identity preservation, temporal consistency, and overall video quality.

%% file: sections/01_introduction.tex
\section{Introduction}\label{sec:intro}

Recent advances in video generative models~\citep{OpenAI2023Sora,yang2024cogvideox,Vidu,Keling,wan2025} have achieved remarkable fidelity video generation, yet their primary reliance on sparse inputs like a single text prompt or initial frame significantly limits their controllability, especially for complex scenes. Drawing inspiration from the compositional flexibility of modern image generators~\citep{deng2025emergingpropertiesunifiedmultimodal,nanobanana}, we focus on the challenging task of multi-subject video generation. This capability is crucial for downstream applications such as subject-driven video synthesis, dynamic scene composition, and controllable product placement, promising to advance personalized content creation, virtual storytelling, and advertising.

However, extending existent single-subject video generation techniques~\citep{he2024id,yuan2024identity,xue2025standinlightweightplugandplayidentity,sang2025lynx} to multiple subjects introduces a significant challenge: maintaining the distinct identity of each subject while ensuring temporal coherent throughout the video. Recent studies~\citep{CINEMA,liu2025phantom} attempt to address this by injecting multi-subject information into pretrained video diffusion models~\citep{wan2025}. But these methods struggle to resolve semantic conflicts between subjects and leads to a degradation of identity. The fundamental challenge lies in the inherent tension between preserving individual subject identities and generating a coherent, dynamic scene—a conflict that current methods fail to resolve effectively.

To address this dilemma, we propose \name{}, which harmonizes identity consistency with motion quality by integrating three strategic components. \name{} is specifically engineered to tackle the inherent challenges of multi-subject video synthesis:
(1) First, to disentangle complex inter-subject relationships while preserving individual attributes, we introduce a \textbf{hierarchical identity-preserving attention} mechanism. This module operates in a cascaded manner: it first attends to intra-subject details to capture unique features, subsequently models inter-subject interactions to mitigate identity leakage, and finally performs cross-modal attention to ensure semantic alignment with the textual prompt. This hierarchical structure establishes a robust foundation for identity preservation.
(2) Second, to enhance generation fidelity for complex descriptions, inspired by prior works~\citep{fang2024vimi,CINEMA}, we utilize a foundational VLM~\citep{Qwen2.5-VL} for \textbf{semantic feature extraction}. unlike conventional text encoders~\citep{radford2021learning}, VLMs offer a granular understanding of scene composition and cross-modal dynamics, which is essential for accurately interpreting and rendering intricate multi-subject scenarios.
(3) Third, to directly optimize the trade-off between identity preservation, video quality, and motion fluidity, we introduce an \textbf{online Reinforcement Learning (RL) strategy}. Optimizing a multi-subject diffusion policy presents significant stability challenges. Inspired by Flow-GRPO~\citep{flowgrpo}, we address this by employing groupwise comparison for advantage estimation~\citep{shao2024deepseekmath,Deepseek-R1}, thereby eliminating the need for a fragile value network. Since applying Online RL (e.g., ReFL~\citep{ReFL} or DRaFT~\citep{DRaFT}) to a multi-subject generative architecture presents non-trivial challenges, as the optimization policy encompasses the video diffusion model augmented with our hierarchical ID attention.  We address this by utilizing groupwise comparison for advantage estimation~\citep{shao2024deepseekmath,Deepseek-R1}, thereby eliminating the reliance on a fragile value network. Furthermore, we introduce a contrastive learning mechanism that provides a stable training signal to fine-tune the hierarchical attention, directly maximizing identity fidelity while mitigating reward hacking.

As illustrated in Fig.~\ref{fig:teaser}, these synergistic components empower \name{} to generate high-fidelity videos distinguished by precise identity preservation, robust prompt alignment, and exceptional temporal consistency. Consequently, our method markedly surpasses previous methods (e.g., Phantom~\citep{liu2025phantom}), in producing subject-consistent videos. Our contributions are summarized as follows:

\begin{itemize}

\item We propose \name{}, a unified framework that mitigates the tension between identity preservation and motion fluidity in multi-subject video generation through the integration of hierarchical attention, VLM-based semantic guidance, and RL-based fine-tuning.

\item We pioneer the application of online RL to multi-subject video generation, demonstrating its effectiveness and stability in optimizing complex perceptual rewards and balancing competing generation objectives.

\item Extensive experiments show the superiority of proposed method over existing methods in terms of identity preservation, temporal consistency, and overall video quality.

\end{itemize}

%% file: sections/02_related_work.tex
\section{Related Works} \label{sec:related_works}

\vspace{-0.2cm}
\paragraph{Subject-Driven Image Generation}
Recent advances in subject-driven image generation have enabled the creation of high-fidelity visual assets, profoundly impacting digital content creation. A prominent line of work involves tuning-based methods, which achieve high subject fidelity by fine-tuning a generative model on a small set of reference images. These approaches range from optimizing the entire model's weights, as in DreamBooth~\citep{ruiz2023dreambooth}, to learning subject-specific textual embeddings~\citep{gal2022image} or adopting parameter-efficient fine-tuning strategies~\citep{hu2021lora,han2023svdiff,yuan2023inserting}. However, a significant drawback of these methods is the substantial computational overhead required for per-subject optimization. To circumvent this limitation, tuning-free methods~\citep{ye2023ip,wang2024instantid,sun2024rectifid,wu2024infinite, liu2025mosaic} have been proposed, which inject identity information during inference via various conditioning mechanisms, thereby offering a more efficient alternative. More recently, a new paradigm has emerged with versatile, unified models~\citep{chen2024unireal, xiao2024omnigen, erwold-2024-qwen2vl-flux, wu2025qwen,seedream,nanobanana} that incorporate subject-driven generation as one of many capabilities. Despite this progress, extending subject-driven generation from the image to the video domain remains a formidable challenge, primarily due to the difficulty of maintaining both identity and temporal consistency across frames.

% More recently, image editing models~\citep{} such as  Gemini Flash Image and SeedDream have enabled natural-language–driven image generation but frequently compromise
% content fidelity, intricate attribute control.

\vspace{-0.2cm}
\paragraph{Subject-Consistent Video Generation}

% Optimization-based subject-to-Video generation, such as Kling~\citep{Keling} require fine-tuning on multiple videos, incurring significant computational costs. 
% Recently, subject-consistent video generation~\citep{Keling,Vidu,Pika,VACE} is commonly achieved by enhancing the model's attention mechanisms to incorporate appearance cues. More flexible paradigms have also emerged, including adapter-based methods like ID-Animator~\citep{he2024id}, and ConsisID~\citep{yuan2024identity}. For the specific challenge of handling multiple subjects, a router mechanism has been introduced~\citep{mou2025dreamo}. Our work is most closely related to Phantom~\citep{liu2025phantom}, Concat-ID~\citep{Concat-ID}, SkyReels-A2~\citep{fei2025skyreels}, and CINEMA~\citep{CINEMA}, which also focus on attention-based feature injection. However, these methods often struggle with conflicts between subject identity and textual prompts, and with maintaining consistency. Moreover, \name{} proposes a novel method to mitigate these conflicts and enforce multi-level consistency, thereby improving the robustness and quality of video generation, particularly in multi-subject scenarios.
Subject-consistent video generation~\citep{Keling,Vidu,Pika,VACE} is commonly achieved by enhancing the model's attention mechanisms to incorporate appearance cues. More flexible paradigms have also emerged, including adapter-based methods like ID-Animator~\citep{he2024id}, CustomVideo~\cite{wang2024customvideo}, ConsisID~\citep{yuan2024identity}, and Stand-In~\citep{xue2025standinlightweightplugandplayidentity}. For the specific challenge of handling multiple subjects, a router mechanism has been introduced~\citep{mou2025dreamo}. Other works have explored richer customization dimensions: \citet{wang2025fantasyportrait} design FantasyPortrait to enhance multi-character portrait animation through expression-augmented diffusion transformers. \citet{hu2025polyvivid} propose PolyVivid to improve vividness and consistency in multi-subject scenarios via cross-modal interaction. \citet{huang2025videomage} present Videomage, which enables both multi-subject and motion customization in text-to-video diffusion. Our work is most closely related to Phantom~\citep{liu2025phantom}, Concat-ID~\citep{Concat-ID}, SkyReels-A2~\citep{fei2025skyreels}, and CINEMA~\citep{CINEMA}, which focus on attention-based feature injection. However, these methods, as well as the above extensions, often struggle with conflicts between subject identity and textual prompts and with maintaining consistency. Moreover, \name{} proposes a novel method to mitigate these conflicts and enforce multi-level consistency, thereby improving the robustness and quality of multi-subject video generation.

% \subsection{Semantic Understanding for Generation}
% A recent trend is the unification of multimodal understanding and generation, with models demonstrating capabilities in both vision and language tasks~\citep{wanx_ace, chen2024unireal, xiao2024omnigen}. Qwen-Image~\citep{} and Waver~\citep{} also show the benefits of integrating the powerful understanding model into generation framework.
% A common strategy for this fusion is to inject visual features from pre-trained encoders or VLMs into a generative backbone using cross-attention mechanisms~\citep{CINEMA}. While powerful, applying this to subject-driven generation, especially with multiple subjects, presents challenges.

\vspace{-0.2cm}
\paragraph{Reinforcement Learning for Generation}
Leveraging its computational efficiency, Direct Preference Optimization (DPO)~\citep{seedream} has become a prevalent technique for post-training text-to-image models to align them with human preferences, such as aesthetics and prompt fidelity. Subsequent work, like DenseDPO~\citep{DenseDPO}, extends this paradigm to the video domain by refining paired data construction and the granularity of preference labels, yielding videos with superior visual quality and motion dynamics. A primary limitation of these methods, however, is their offline nature, which precludes online parameter updates. This challenge has been recently addressed by verifiable reward-based models like OpenAI’s o1~\citep{OpenAI-O1} and Deepseek-R1~\citep{Deepseek-R1,shao2024deepseekmath}. Concurrently, the growing interest in online learning has spurred various online Reinforcement Learning (RL) methods, including ReFL~\citep{ReFL}, RAFT~\cite{raft}, DRaFT~\citep{DRaFT}, FlowGRPO~\citep{flowgrpo}, DanceGRPO~\citep{dancegrpo},  SRPO~\citep{SRPO} and Identity-GRPO~\citep{meng2025identity}, which have demonstrated promising results in enhancing  aesthetics, text-rendering fidelity and face similarity. 
Nevertheless, their application to subject-consistent video generation remains limited. To bridge this gap, we introduce task-specific rewards tailored for subject-consistent video generation, a method that mitigates reward hacking and delivers videos with high identity consistency.

%% file: sections/03_method.tex
\begin{figure*}[!t]
    \centering
    \includegraphics[width=\linewidth]{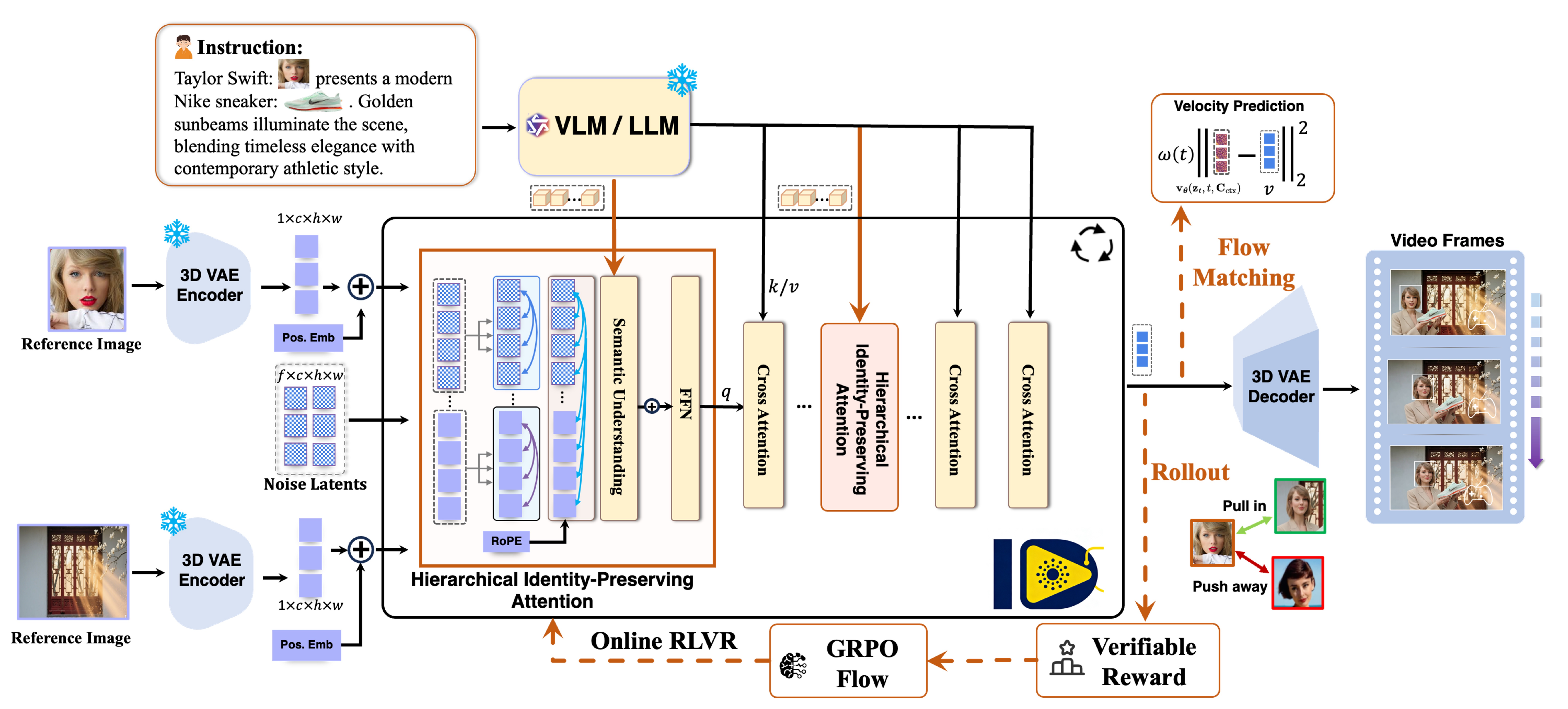}
    \caption{\textbf{\name{} Overview}. Our model incorporates a hierarchical identity-preserving attention mechanism and a VLM that performs reasoning on the multimodal input into a video DiT to enable multi-subject video generation. An online RL stage further refines the concept alignment.}
    \label{fig:overview}
\end{figure*}

\vspace{-0.2cm}
\section{Method}\label{sec:method}
We are addressing the task of multi-subject video generation. Given a textual prompt $\mathbf{C}_{\text{txt}}$ and a set of $N$ reference images $\mathcal{I} := \{\mathbf{I}_k\}_{k=1}^N$, where each $\mathbf{I}_k$ depicts a unique subject, the objective is to generate a video $\mathbf{V}$ that aligns with the prompt $\mathbf{C}_{\text{txt}}$ and faithfully preserves the identities of all $N$ subjects from $\mathcal{I}$ with high fidelity and temporal coherence. In the following sections, we first delineate the preliminaries of our method (Sec.~\ref{sec:Preliminaries}) and subsequently introduce the architecture of the proposed method (Sec.~\ref{sec:our_method}), training schemes (Sec.~\ref{sec:training}), and the data curation pipeline (Sec.~\ref{sec:data}). 

\subsection{Rectified Flow for Video Generation} \label{sec:Preliminaries}

Our methodology is founded upon a latent video diffusion transformer~\citep{peebles2023scalable,wan2025} trained with standard Rectified Flow~(RF)~\citep{lipman2022flow, liu2022flow, esser2024scaling}. In latent video diffusion, a video clip $\mathbf{V} \in \mathbb{R}^{T \times H \times W \times 3}$ is first projected into a latent space by the encoder $\mathcal{E}$ of a variational autoencoder (VAE), yielding the representation $\mathbf{z}_0 := \mathcal{E}(\mathbf{V}) \in \mathbb{R}^{f \times c \times h \times w}$. The corresponding VAE decoder is denoted by $\mathcal{D}(\mathbf{z}_0)$. The generative process is formulated based on Rectified Flow, which defines a straight-line trajectory between a sample from the data distribution $\mathbf{z}_0 \sim p_{\text{data}}$ and a sample from a simple prior distribution, typically a standard Gaussian $\boldsymbol{\epsilon} \sim \mathcal{N}(\mathbf{0}, \mathbf{I})$. The trajectory is parameterized by a time variable $t \in [0, 1]$:
\begin{equation}
    \mathbf{z}_t = (1-t) \mathbf{z}_0 + t\boldsymbol{\epsilon}.
\end{equation}
This path induces a constant velocity field $\mathbf{v} = \boldsymbol{\epsilon} - \mathbf{z}_0$. The core objective is to train a video diffusion transformer $\mathbf{v}_{\boldsymbol{\theta}}$, parameterized by $\boldsymbol{\theta}$, to predict this velocity. The training objective is formulated as a regression loss over this velocity field:
\begin{equation}\label{eq:loss}
    \mathcal{L}_{\text{RF}} = \mathbb{E}_{t, \mathbf{z}_0, \boldsymbol{\epsilon}} \left[ w(t) \left\lVert \mathbf{v}_{\boldsymbol{\theta}}(\mathbf{z}_t, t, \mathbf{C}_{\text{ctx}}) - (\boldsymbol{\epsilon} - \mathbf{z}_0) \right\rVert_2^2 \right],
\end{equation}
where $\mathbf{C}_{\text{ctx}}$ represents the conditioning information (e.g., a text or image prompt), and $w(t)$ is a weighting function that balances the loss across different noise levels.

\subsection{\textbf{\name{}}: Multi-Subject Video Generation}\label{sec:our_method}
As shown in Fig.~\ref{fig:overview}, \name{} extends the DiT-based latent video diffusion architecture by introducing two key architectural innovations to handle the complexities of multi-subject conditioning: (1) A hierarchical identity-preserving attention mechanism, which is designed to hierarchically aggregate features. 
(2) Semantic understanding via a pretrained VLM, which we employ as a sophisticated encoder for both textual and visual inputs. 
To further incentivize the model's semantic alignment and identity preservation, we incorporate an online reinforcement learning phase subsequent to the initial flow matching training. 

\paragraph{Hierarchical Attention and Semantic Grounding.}
Our model is based on a Wan Video~\citep{wan2025}, where latent video tokens and conditioning tokens are processed together. We first encode the input reference images $\mathcal{I}$ using a pretrained image encoder to obtain visual feature maps $\{\mathbf{F}_k\}_{k=1}^N$, where each $\mathbf{F}_k \in \mathbb{R}^{c \times h \times w}$ corresponds to the $k$-th subject. These feature maps are then flattened into token sequences $\{\mathbf{f}_k\}_{k=1}^N$, with each $\mathbf{f}_k \in \mathbb{R}^{hw \times c}$. To effectively integrate information from multiple subjects and modalities, we introduce a hierarchical identity-preserving attention mechanism. This operates in three stages: (1) Intra-subject attention captures fine-grained details within each subject; (2) Inter-subject attention models the interactions between different subjects; and (3) Multi-modal attention fuses subject features with text and video tokens. This hierarchical design is crucial for preserving individual identities while capturing complex inter-subject dynamics.

Recently, VLMs have seen increasing adoption in modern video generation models~\citep{gao2025seedance,zhang2025waver}, yet their potential for \textit{disentangling} complex multi-subject dynamics remains significantly underutilized. Notably, no prior open-source work has integrated a VLM as the core reasoning engine within the powerful open-source Wan-Video architecture. Our approach fundamentally differs by harnessing the VLM's granular cross-modal reasoning to actively guide our Hierarchical Identity-Preserving Attention. This transforms the VLM from a static encoder into a dynamic semantic guide. 
We employ the Qwen2.5-VL model~\citep{Qwen2.5-VL} to process the textual prompt $\mathbf{C}_{\text{txt}}$ and reference images $\mathcal{I}$, yielding semantically enriched tokens: $\mathbf{f}_{\text{txt}} = \text{VLM}_{\text{enc}}(\mathbf{C}_{\text{txt}}, \mathcal{I}) \in \mathbb{R}^{l' \times c}$. This structured representation provides an explicit signal to the generative process. The complete set of conditioning tokens $\mathbf{C}_{\text{ctx}}$ is subsequently formed by concatenating these enhanced text tokens with the individual subject tokens: $ \mathbf{C}_{\text{ctx}} = [\mathbf{f}_{\text{txt}}; \mathbf{f}_1; \ldots; \mathbf{f}_N] \in \mathbb{R}^{(l' + N \cdot hw) \times c}. $ Our central innovation is thus bridging the VLM's high-level semantic reasoning with the generative model's low-level spatial layout control.

\subsection{Post-training with Online RL}\label{sec:training}
Post-training the base model with a perceptual reward for identity preservation presents a formidable challenge, primarily due to the credit assignment problem. \textit{How to adjust the intricate computations within the hierarchical ID attention layers to enhance a holistic, video-level reward?} While DPO-style methods~\citep{DenseDPO,DPO} offer computational efficiency, their offline nature precludes the online parameter updates essential for this fine-tuning task. Alternatively, standard policy-gradient algorithms~\cite{ReFL,DRaFT} are often impractical in this context. They necessitate training a separate value function to estimate rewards, which is unstable and sensitive to hyperparameters.
Following recent advances~\citep{flowgrpo,dancegrpo}, for each condition $q$, \name{} samples a group of outputs $\{o_1, o_2, \dots, o_G\}$ from the old policy $\pi_{\theta_{old}}$ and then optimizes the policy model by maximizing the following objective: 
\begin{equation}
\begin{aligned}
& \mathcal{J}_{GRPO}(\theta) = \mathbb{E}_{\substack{q \sim P(Q), \\ \{o_i\}_{i=1}^G \sim \pi_{\theta_{\text{old}}}(O|q)}} \frac{1}{G} \sum_{i=1}^G \frac{1}{|o_i|} \sum_{t=1}^{|o_i|} \\
& \quad \left\{ \min \left[ r_t^i(\theta) \hat{A}_{i,t}, \text{clip}(r_t^i(\theta), 1-\epsilon, 1+\epsilon) \hat{A}_{i,t} \right] \right. \\
& \quad \left. - \beta \mathbb{D}_{\text{KL}} \big[\pi_\theta \,\|\, \pi_{\text{ref}}\big] \right\}, \\
& \text{where} \quad r_t^i(\theta) = \frac{\pi_\theta(o_{i,t} \mid q, o_{i,<t})}{\pi_{\theta_{\text{old}}}(o_{i,t} \mid q, o_{i,<t})}.
\end{aligned}
\end{equation} the relative quality $ \hat A_{i, t} $ of the i-th response is computed as $
\hat A_{i, t} = \frac{r_i - \text{Mean}(\{r_1, r_2, \ldots, r_n\})}{\text{Std}(\{r_1, r_2, \ldots, r_n\})},
$

% \begin{equation*}
% \hat A_{i, t} = \frac{r_i - \text{Mean}(\{r_1, r_2, \ldots, r_n\})}{\text{Std}(\{r_1, r_2, \ldots, r_n\})},
% \end{equation*}
 While convert flow matching's determinsitic generation process to a stochastic one by injecting noise at each step of the flow integration. The update rule is defined as: 
\newcommand{\xv}{\mathbf{x}}
\newcommand{\thetav}{\bm{\theta}}

\begin{equation}
\begin{aligned}
\xv_{t+\Delta t} = \xv_t &+ \left[ \vv_{\thetav} (\xv_t, t) + \frac{\sigma_t^2}{2t}\left(\xv_t + (1-t)\vv_\theta(\xv_t, t)\right) \right] \Delta t \\
& + \sigma_t \sqrt{\Delta t}\,\bm{\epsilon}
\end{aligned}
\end{equation}
where $\bm{\epsilon} \sim \mathcal{N}(0, I)$ injects stochasticity into the generation process and $\sigma_t = a \sqrt{\frac{t}{1-t}}$. 

\paragraph{Reward Design.}  The objective is to maximize a composite reward function $\mathcal{R}_{\text{total}}$, defined as a weighted sum of video quality and identity consistency:
\begin{equation}
    \mathcal{R}_{\text{total}}(\mathbf{V}) := w_{\text{fid}} \mathcal{R}_{\text{fid}}(\mathbf{V}, \mathcal{I}) + w_{\text{qual}} \mathcal{R}_{\text{qual}}(\mathbf{V}) ,
\end{equation}
where we empirically set the weights to $w_{\text{fid}}=0.6$, $w_{\text{qual}}=0.4$ to prioritize identity preservation while ensuring high quality and text alignment.

\paragraph{Fidelity Reward ($\mathcal{R}_{\text{fid}}$).}
The fidelity reward ensures that the generated subjects are consistent with the reference images $\mathcal{I}$. It combines a face-specific metric with a holistic subject consistency score:
\begin{equation}
    \mathcal{R}_{\text{fid}} := (1-\alpha) \mathcal{R}_{\text{face}} + \alpha \mathcal{R}_{\text{subject}}.
\end{equation}
For facial identity $\mathcal{R}_{\text{face}}$, We use a robust combination of the mean and minimum FaceSim scores~\citep{consisid} across all $N$ subjects:
\begin{equation}
    \mathcal{R}_{\text{face}} := (1-\gamma) \left( \frac{1}{N}\sum_{k=1}^N \mathcal{R}_{\text{id}}^k \right) + \gamma \left( \min_{k=1,\dots,N} \mathcal{R}_{\text{id}}^k \right),
\end{equation}
where $\mathcal{R}_{\text{id}}^k$ is the ArcFace-based score for subject $k$, and we set $\gamma=0.5$.
For overall subject consistency, $\mathcal{R}_{\text{subject}}$, we employ the NexusScore, a metric from our proposed OpenS2V-Eval benchmark that assesses non-facial and holistic subject attributes. We balance face and subject with $\alpha=0.5$.

\paragraph{Quality Reward ($\mathcal{R}_{\text{qual}}$).}
The quality reward assesses both the aesthetic appeal and the physical plausibility of the generated video $\mathbf{V}$:
\begin{equation}
    \mathcal{R}_{\text{qual}} := (1-\beta) \mathcal{R}_{\text{aes}} + \beta \mathcal{R}_{\text{nat}}.
\end{equation}
$\mathcal{R}_{\text{aes}}$ is the standard Aesthetic Score~\citep{improved-aesthetic-predictor}. To combat reward hacking that may produce unnatural but aesthetically pleasing artifacts, we introduce $\mathcal{R}_{\text{nat}}$. This is the NaturalScore from our benchmark, which leverages a powerful VLM to evaluate whether the video adheres to physical laws and common sense. We set $\beta=0.4$ to strongly penalize unnatural generations.

\begin{figure}[!t]
    \centering
    \includegraphics[width=\linewidth]{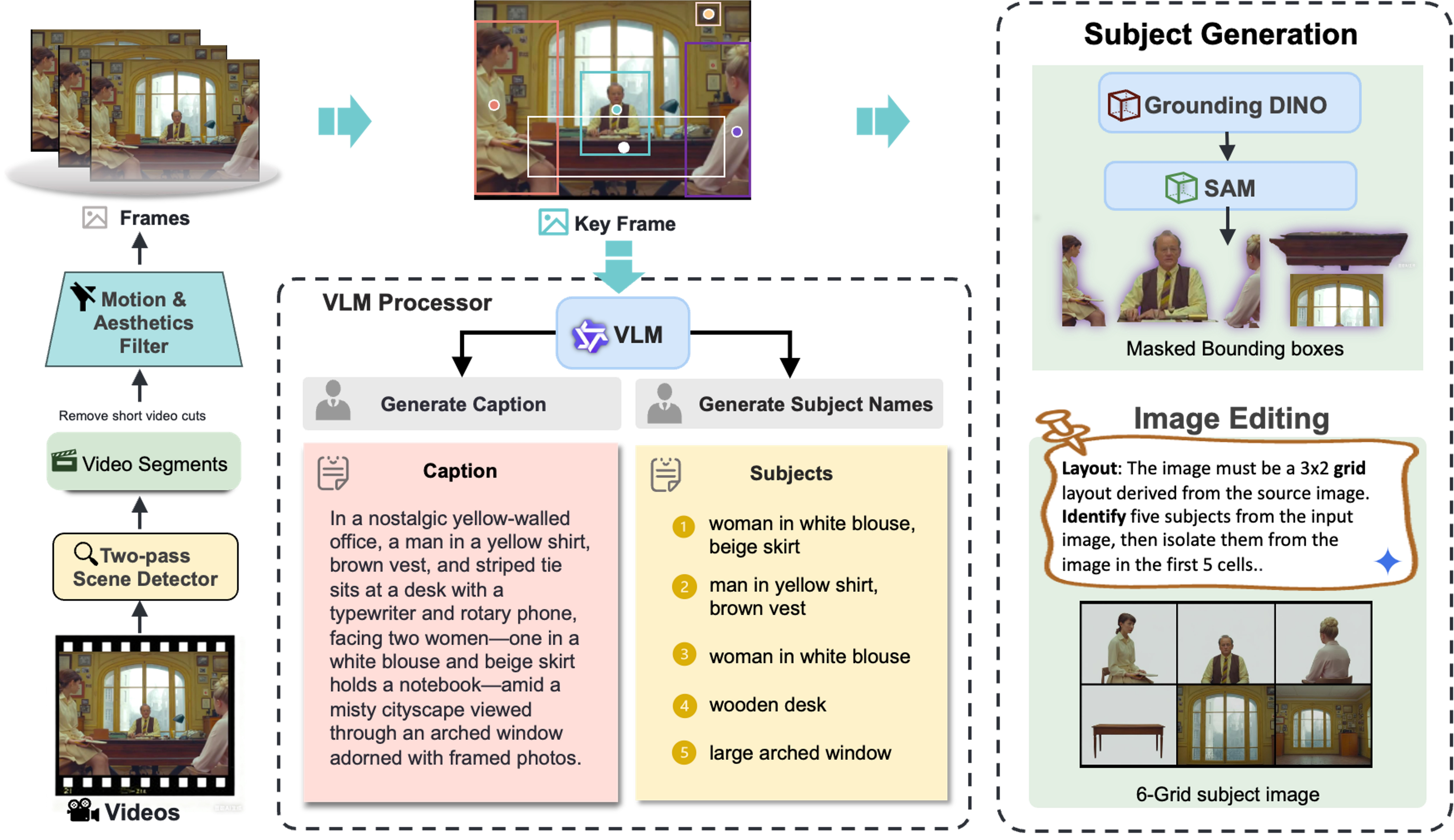}
    \caption{Data curation pipeline of \name{}.  }
    \label{fig:data}
\end{figure}
\subsection{Training Data curation} \label{sec:data}
Current subject-to-video generation methods are hampered by notable deficiencies in output quality and diversity. These shortcomings are largely attributable to the restrictive nature of paired training data, which proves inadequate for modeling complex, real-world variations in subject motion, camera viewpoint, and scene layout. To surmount these limitations, we propose a sophisticated data-curation pipeline that harnesses the capabilities of modern VLMs (e.g., QwenVL-72B)~\citep{Qwen2.5-VL} and a powerful image editing model (a.k.a., Nano Banana)~\citep{nanobanana}. 
 The first component consists of subject-video pairs extracted from a diverse collection of OpenS2V-Nexus ~\citep{yuan2025opens2v}, offering a wide array of authentic scenes and actions. The second component comprises synthetically generated data, where subjects rendered by cutting-edge image editing models are placed into novel contexts, including explicitly designed cross-subject compositions and fusion examples, to mitigate ``copy-plate'' artifacts and enhance multi-entity interaction coherence. The third component is a high-fidelity collection of professionally shot videos with detailed annotations. As shown in Fig.~\ref{fig:data}, the culmination of this pipeline is a large-scale, composite dataset meticulously constructed to propel the development of next-generation multi-subject video synthesis models. This dataset is composed of three heterogeneous data sources. More technical details are provided in Appendix.

%% file: sections/04_experiments.tex
\vspace{-0.2cm}
\section{Experiments}\label{sec:exp}
\subsection{Experimental Settings} \label{sec:implement}
\input{tables/opens2v.tex}

\paragraph{Implementation Details}
Our model, \name{}, is initialized from the weights of the Wan-Video-1.3 B model~\citep{wan2025}. To leverage superior semantic understanding, we employ a dual text-encoder architecture, combining T5~\citep{T5} with Qwen2.5-VL-7B-Instruct~\citep{Qwen2.5-VL}. The model was trained for 30,000 iterations on our curated dataset at a resolution of 480p, utilizing 16 H20 GPUs. For inference, we employ an Euler sampler with 50 steps and classifier-free guidance~\citep{ho2022classifier} to modulate the influence of image and text conditions, setting the classifier-free guidance scale to 2.5. The inference time for our 1.3B model is approximately one minute to generate a video at 480p resolution.
\paragraph{Baselines}
We benchmark our method against state-of-the-art open-source models that support native subject-to-video generation. This includes Phantom~\citep{liu2025phantom} and SkyReels-A2~\citep{fei2025skyreels}, both available in 1.3B and 14B parameter variants, and VACE~\citep{VACE}. For a comprehensive assessment, we also present a qualitative comparison with leading proprietary systems, such as VIDU~\citep{Vidu}, Pika~\citep{Pika} and Kling~\citep{Keling}.

\paragraph{Evaluation Metrics}
Following the benchmark protocol of OpenS2V-Nexus~\citep{yuan2025opens2v}, we perform evaluation on a diverse dataset of \textbf{180 unique subject-text pairs}. To ensure a rigorous and unbiased evaluation, this dataset is maintained as a held-out set, completely disjoint from our training data, thereby preventing any potential data leakage. We report on a suite of automated metrics, including Aesthetics, Motion quality, Face Similarity (FaceSim), and alignment scores against three benchmarks (GmeScore, NexusScore, NaturalScore). In our ablation studies, we also report the VLM-based visual scorer, Q-Align~\cite{wu2023qalign}, for perceptual quality.

\subsection{Main Results}

\paragraph{Quantitative Comparison}
As demonstrated in Table~\ref{tab:subject-to-video-custom}, \name{} achieves state-of-the-art performance across all evaluation metrics, outperforming existing open-source models by a significant margin. Notably, our method shows substantial improvements in FaceSim and NexusScore, which we attribute to our hierarchical identity-preserving attention mechanism and the rich semantic guidance from the VLM encoder. These components work in concert to ensure high-fidelity identity preservation and robust alignment with complex textual prompts.

\begin{figure*}[t]
    \centering
    \includegraphics[width=0.95\linewidth]{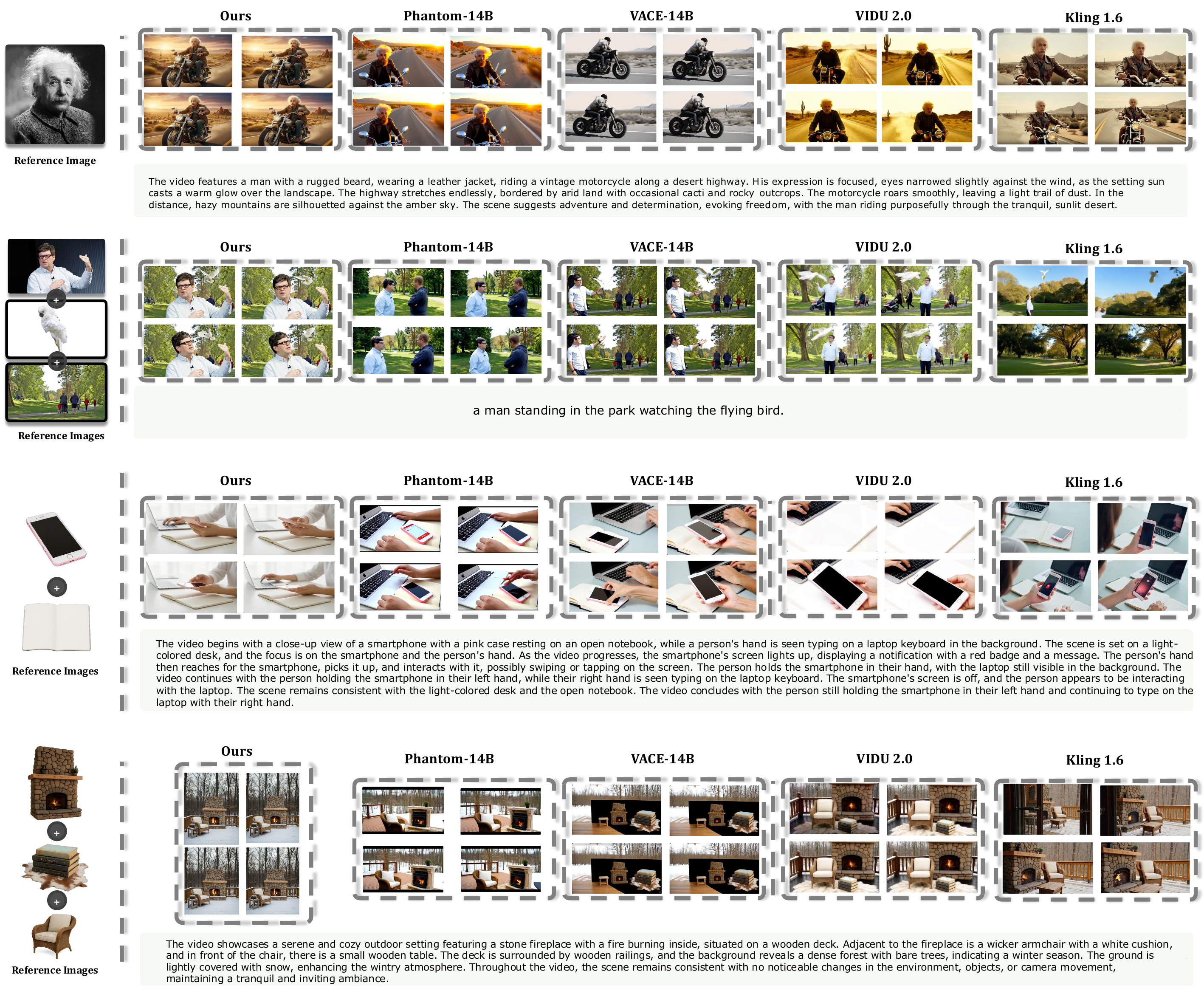}
    \caption{\textbf{Qualitative Comparison with State-of-the-Art Methods.} Our method, \name{}, demonstrates superior performance in identity preservation, temporal consistency, and alignment with the textual prompt compared to existing open-source and proprietary models. }
    \label{fig:main_comparison}
    \vspace{-0.4cm}
\end{figure*}

\vspace{-0.2cm}
\paragraph{Qualitative Comparison}
Fig.~\ref{fig:main_comparison} presents a qualitative comparison against leading methods. \name{} consistently generates videos with superior visual quality, temporal coherence, and identity preservation. While baseline methods often exhibit artifacts or identity drift, especially in longer videos or complex scenes, our model maintains a stable representation of all subjects , accurately reflecting their appearance from the reference images.

% Additional visualizations are available in Appendix Fig.~\ref{fig:main_comparison}.

\subsection{Ablation and Analysis}
% We conduct a series of ablation studies to validate the effectiveness of our key design choices. The results, summarized in Table~\ref{tab:ablation}, underscore the importance of each proposed component.
% \vspace{-0.2cm}
\paragraph{Effect of Hierarchical ID-Preserving Attention}
To isolate the contribution of our attention mechanism, we replaced it with a standard cross-attention mechanism that simply concatenates all subject and text tokens. As shown in Table~\ref{tab:ablation}, this variant (``w/o Hierarchical Attention'') suffers a significant drop in FaceSim, confirming that our hierarchical approach is crucial for modeling subject-specific features and preventing identity degradation.

\vspace{-0.4cm}
\paragraph{Effect of VLM-based Semantic Understanding}
We evaluated the impact of the VLM by replacing the dual-encoder setup with a single T5 encoder~\citep{T5}. The results (``w/o VLM Encoder'') show a marked decrease in Text-Video Alignment. This highlights the VLM's superior ability to parse complex, multi-subject prompts and provide fine-grained semantic guidance, which is essential for accurate video generation.

\begin{figure*}[!t]
    \centering
    \includegraphics[width=0.9\linewidth]{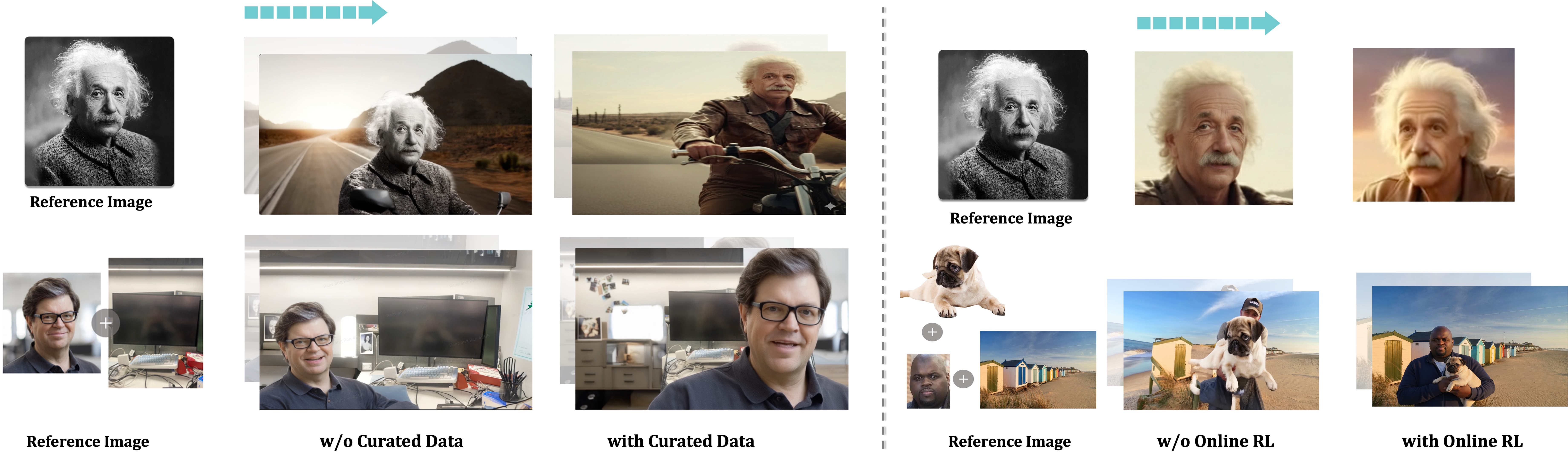}
    \caption{\textbf{Qualitative Ablation Study.} The left panel highlights the importance of our curated dataset, demonstrating improved coherence and realism in subject integration (e.g., mitigating `copy-paste' artifacts) compared to a model trained without it. Meanwhile, the right panel illustrates the effectiveness of our online reinforcement learning stage, which significantly enhances visual quality and subject consistency. }
    \label{fig:ablation_qualitative}
\end{figure*}

\vspace{-0.2cm}
\paragraph{Effect of Data Curation Pipeline}
Finally, we trained the full model on a baseline dataset without our curated data. This version (``w/o Curated Data'') yielded lower scores across all metrics, particularly in Video Quality. Critically, the base model exhibited more severe ``copy-plate'' artifacts, due to limited exposure to cross-subject interactions. Our data curation pipeline explicitly mitigates this by incorporating synthetic cross-subject compositions and fusion examples, which enhances the model's ability to synthesize coherent multi-entity interactions. This demonstrates that our data curation pipeline, which synthesizes diverse and high-quality training examples, is vital for achieving robust and generalizable performance. Qualitative results in Fig.~\ref{fig:ablation_qualitative} further illustrate these findings.

\input{tables/ablation.tex}

\begin{figure}[!t]
    \centering
    \includegraphics[width=\linewidth]{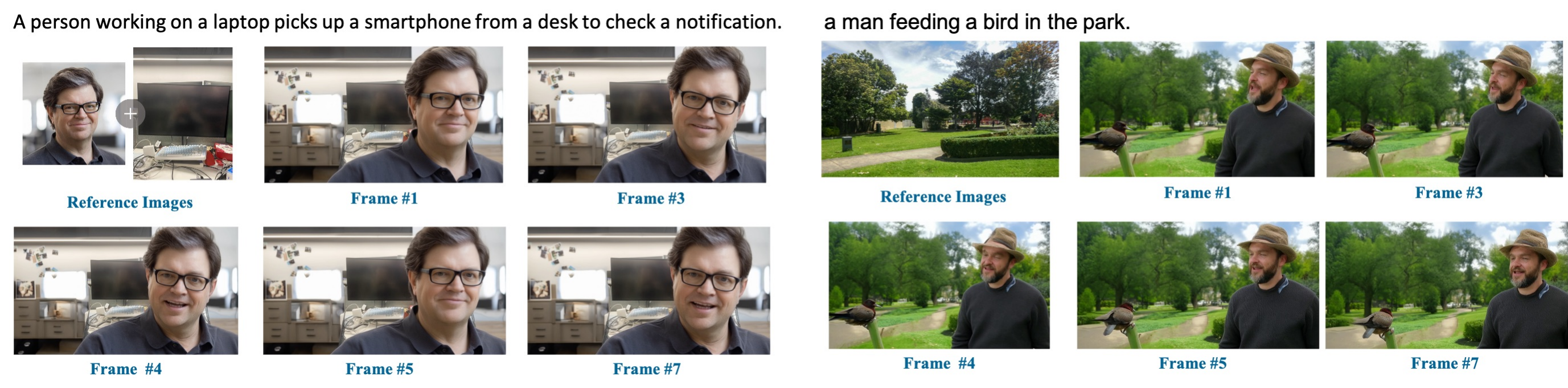}
    \caption{\textbf{Applications in Controllable Video Editing.} \name{} enables zero-shot editing of existing videos, including subject replacement and background modification, while preserving identity and temporal consistency. }
    \label{fig:applications}
\end{figure}
% To validate the effectiveness of our online reinforcement learning stage, we compare the full model with a variant trained only with standard SFT.
\paragraph{Effect of Online RL}
 As shown in Tab.~\ref{tab:rl_ablation}, the online RL stage brings substantial improvements across all metrics compared to the SFT baseline. Specifically, our model achieves a 13.7\% relative increase in FaceSim  and a 14.9\% relative increase in Aesthetics. The alignment score, Q-Align, also sees a significant 16.8\% relative improvement. This demonstrates that directly optimizing the model with our composite reward function is crucial for enhancing identity preservation and perceptual quality. The qualitative results in Fig.~\ref{fig:ablation_qualitative} (right panel) further corroborate this finding, where the model trained with RL generates videos with significantly better subject consistency compared to the SFT baseline.
\vspace{-0.4cm}
 \input{tables/rl_ablation.tex}
\paragraph{Analysis of Online RL }
 Offline RL methods like DPO~\citep{DPO} offer an alternative but rely on a static dataset of preference pairs, which may not sufficiently cover the vast space of possible generations. In contrast, our online approach allows the policy to actively explore and receive feedback from our reward model, leading to more effective optimization. As shown in Tab.~\ref{ ↑:rl_ablation}, our online GRPO method significantly outperforms both the SFT baseline and the offline approach across all metrics. This confirms that online sample generation is more effective for fine-tuning video generation models. The table also presents an ablation study on the components of our composite reward function, $\mathcal{R}_{\text{total}}$. Removing the fidelity term ($\mathcal{R}_{\text{fid}}$) causes a severe drop in FaceSim, while removing the quality ($\mathcal{R}_{\text{qual}}$) term leads to a degradation in Aesthetics. Interestingly, removing the Natural Reward $\mathcal{R}_{\text{nat}}$ leads to a degradation in Q-Align, but a slight increase in FaceSim and Aesthetics. This suggests a reward hacking scenario, where the model over-optimizes for certain metrics at the expense of overall alignment. These results validate our multi-faceted reward design and the necessity of each component for achieving a balanced improvement.
 \vspace{-0.4cm}

\paragraph{Controllable Video Editing}  The architecture of \name{} naturally lends itself to a range of editing applications. By providing an existing video as an initial state and conditioning the generation on new subjects or modified text prompts, our model can perform zero-shot, identity-consistent video editing. As illustrated in Fig.~\ref{fig:applications}, \name{} can seamlessly insert new subjects into a scene, replace existing ones, or alter the background, all while maintaining temporal and semantic coherence. This capability opens up new avenues for personalized content creation and virtual storytelling.

%% file: tables/opens2v.tex
\begin{table*}[!t]
\centering
\renewcommand\arraystretch{1.2}
\resizebox{0.95 \textwidth}{!}{
\begin{tabular}{lccccccc}
\toprule[1.2pt]
Method & Total Score$\uparrow$ & Aesthetics$\uparrow$ & Motion$\uparrow$ & FaceSim$\uparrow$ & GmeScore$\uparrow$ & NexusScore$\uparrow$ & NaturalScore$\uparrow$ \\
\midrule
\multicolumn{8}{l}{\textbf{\textit{Proprietary Models}}} \\
% \midrule
Vidu 2.0~\citep{Vidu} & 47.59\% & 41.47\% & 13.52\% & 35.11\% & 67.57\% & 43.55\% & 71.44\% \\
Pika 2.1~\citep{Pika} & 48.88\% & 46.87\% & 24.70\% & 30.80\% & 69.21\% & 45.41\% & 69.79\% \\
Kling 1.6~\citep{Keling} & 54.46\% & 44.60\% & 41.60\% & 40.10\% & 66.20\% & 45.92\% & 79.06\% \\
% Dreamina~\citep{dreamina} & \\
\midrule[1.2pt]
\multicolumn{8}{l}{\textbf{\textit{Open-source Models}}} \\
% \midrule
VACE-1.3B~\citep{VACE} & 45.53\% & 48.24\% & 18.83\% & 20.58\% & 71.26\% & 37.95\% & 71.78\% \\
VACE-14B~\citep{VACE} & 52.87\% & 47.21\% & 15.02\% & 55.09\% & 73.27\% & 44.20\% & 72.78\% \\
Phantom-1.3B~\citep{liu2025phantom} & 50.71\% & 46.67\% & 14.29\% & 48.55\% & 69.43\% & 42.44\% & 70.26\% \\
Phantom-14B~\citep{liu2025phantom} & 52.32\% & 46.39\% & 33.42\% & 51.48\% & 70.65\% & 37.43\% & 68.66\% \\
SkyReels-A2-P14B~\citep{fei2025skyreels} & 49.61\% & 39.40\% & 25.60\% & 45.95\% & 64.54\% & 43.77\% & 67.22\% \\
\midrule

\textbf{Ours-1.3B (Base Model)} \scalebox{1.5}{\color{sh_red}{$\star$}} & 54.33\% & 42.50\% & 38.00\% & 58.12\% & 64.06 \% & 43.22\% & 71.09\% \\
\textbf{Ours-1.3B (Base Model + online RL)} \scalebox{1.5}{\color{sh_blue}{$\star$}} & 55.16\% & 48.85\% & 36.50\% & 66.10\% & 62.20\% & 43.45\% & 69.15\% \\
\textbf{Ours-14B (Large-scale Model)} & 57.05\% & 45.28\% & 40.34\% & 60.71\% & 67.49\% & 45.11\% & 73.23\% \\
\bottomrule[1.2pt]
\end{tabular}
}
\caption{\textbf{Quantitative Comparison against existing methods for the open-domain subject-to-video benchmark}. Total score is the normalized weighted sum of other scores. ``$\uparrow$'' indicates that higher is better. }
\label{tab:subject-to-video-custom} 
\end{table*}

%% file: tables/ablation.tex
\begin{table}[t]
\centering
\caption{\textbf{Ablation studies on the key components of \name{}.} We compare the full model against variants removing hierarchical attention, the VLM encoder, and the curated dataset. }
\label{tab:ablation}

\newcommand{\decline}[1]{\textcolor{gray}{\scriptsize{($\downarrow$#1)}}}

\resizebox{\linewidth}{!}{%
\begin{tabular}{lcccc}
\toprule[1.2pt]
Method & FaceSim $\uparrow$ & Q-Align $\uparrow$ & Video Quality $\uparrow$ & Total Score $\uparrow$ \\
\midrule

\rowcolor{gray!10} 
\textbf{Ours-1.3B (Base Model)} \scalebox{1.5}{\color{sh_red}{$\star$}} & \textbf{58.12\%} & \textbf{0.351} & \textbf{48.91\%} & \textbf{54.33\%} \\
\midrule

\quad w/o Hierarchical Attention & 51.34\% \decline{11.7\%} & 0.348 \decline{0.9\%} & 47.52\% \decline{2.8\%} & 50.11\% \decline{7.8\%} \\

\quad w/o VLM Encoder & 56.98\% \decline{2.0\%} & 0.287 \decline{18.2\%} & 46.88\% \decline{4.2\%} & 49.89\% \decline{8.2\%} \\

\quad w/o Curated Data & 54.55\% \decline{6.1\%} & 0.321 \decline{8.5\%} & 45.13\% \decline{7.7\%} & 48.78\% \decline{10.2\%} \\
\bottomrule[1.2pt]
\end{tabular}%
}
\vspace{-0.4cm}
\end{table}

%% file: tables/rl_ablation.tex
\begin{table}[h]
\centering
\caption{Analysis of the online RL optimization. We compare our online GRPO approach with offline DPO and a standard SFT baseline. We also ablate the components of our composite reward function $\mathcal{R}_{\text{total}}$.}
\label{tab:rl_ablation}
\resizebox{\linewidth}{!}{%
\begin{tabular}{l|cccc}
\toprule[1.2pt]
Method & FaceSim $\uparrow$ & Aesthetics $\uparrow$ & Q-Align. $\uparrow$ & Total Score $\uparrow$ \\
\midrule
SFT Baseline & 58.12\% & 42.50\% & 0.351  & 54.33\% \\
\midrule
\multicolumn{5}{l}{\textit{Comparison of RL Algorithms}} \\
DPO (Offline) & 62.35\% & 45.15\% & 0.382 & 54.80\% \\
Ours (Online) & \textbf{66.10\%} & \textbf{48.85\%} & \textbf{0.410} &  \textbf{55.16\%} \\
\midrule
\multicolumn{5}{l}{\textit{Ablation on Reward Components}} \\
Ours w/o Fidelity ($\mathcal{R}_{\text{fid}}$) & 45.32\% & 47.10\% & 0.391 & 53.50\% \\
Ours w/o Quality ($\mathcal{R}_{\text{qual}}$) & 63.50\% & 43.81\% & 0.379 & 54.82\% \\
Ours w/o Natural ($\mathcal{R}_{\text{nat}}$) & \textbf{69.30\%} & \textbf{50.83\%} & 0.361 & 53.01\% \\
Ours (Online) & 66.10\% & 48.85\% & \textbf{0.410} &  \textbf{55.16\%} \\
\bottomrule[1.2pt]
\end{tabular}
}
\vspace{-0.4cm}
\end{table}

%% file: sections/05_conclusion.tex
\vspace{0.2cm}
\section{Conclusion}\label{sec:conclu}

% In this work, we presented \name{}, a novel framework designed to address the challenges of multi-subject video generation. Our approach integrates two key innovations into a latent video diffusion model: a hierarchical identity-preserving attention mechanism and a powerful pretrained Vision-Language Model for semantic understanding. By moving beyond simple feature concatenation, \name{} effectively models complex interactions within and between subjects, as well as across modalities. This design allows the model to preserve subject identities, integrate semantic information, and maintain temporal consistency with high fidelity. Extensive experiments demonstrate that \name{} achieves state-of-the-art performance, outperforming existing methods across a comprehensive suite of metrics. Our work marks an important step towards general and scalable multi-subject video generation in complex, real-world scenarios.

In this work, we introduced \name{}, an innovative framework for multi-subject video generation that integrates hierarchical identity-preserving attention and a VLM into a latent diffusion backbone. Moving beyond simple feature concatenation, \name{} captures complex intra- and inter-subject interactions, preserves identities, incorporates semantics, and ensures temporal consistency. Extensive experiments demonstrate that \name{} achieves state-of-the-art performance, outperforming existing methods across comprehensive metrics. Our work marks an important step towards general and scalable multi-subject video generation in complex, real-world scenarios. 
\textbf{Limitations and Future Work} 
Despite its strength in multi-subject identity preservation, \name{} has limitations in modeling complex interactions and fine-grained dynamics. Future work will focus on incorporating physics-aware priors, mitigating biases in pretrained components, and advancing fine-grained controllable generation of attributes, actions, and interactions.

%% file: sections/99_appendix.tex
\section*{Appendices Overview}
\addcontentsline{toc}{section}{Appendices Overview}
The supplementary material includes the following sections:

\begin{itemize}
    \item Sec.~\ref{sec:implement}: More Implementation Details.
    \begin{itemize}
        \item Details of Network Architecture (Sec.~\ref{apx:network_arch}).
        \item Details of Hierarchical Identity-Preserving Attention (Sec.~\ref{apx:attention}).
        \item Details of Reward Calculation (Sec.~\ref{reward}).
    \end{itemize}

    \item Sec.~\ref{apx:dataset}: Dataset Curation Details.
    \begin{itemize}
        \item Dataset Composition and Statistics (Sec.~\ref{apx:dataset_stats}).
        \item Reference Image Construction Strategies (Sec.~\ref{apx:data_comp}).
    \end{itemize}
    
    \item Sec.~\ref{apx:metric}: Evaluation Metric Details.

    \item Sec.~\ref{apx:results}: Additional Results and Analysis.
    \begin{itemize}
        \item Evaluation on Multi-Subject Scenarios (Sec.~\ref{apx:multi_subject}).
        \item Human Preference Study (Sec.~\ref{apx:human_study}).
    \end{itemize}

    \item Sec.~\ref{apx:discussion}: Additional Discussion.
    \begin{itemize}
        \item Computational Overhead (Sec.~\ref{apx:computation}).
        \item Ethics Statement (Sec.~\ref{apx:ethics}).
    \end{itemize}
\end{itemize}

\section{More Implementation Details} \label{sec:implement}
\input{tables/arch}
\input{tables/attention_ablation}

\subsection{Network Architecture} \label{apx:network_arch}
Our model's architecture is founded on three synergistic modules: a Vision-Language Model (\textbf{VLM}), specifically the Qwen2.5-VL 7B variant~\citep{Qwen2.5-VL}, for advanced multi-modal comprehension; a Variational Autoencoder (\textbf{VAE})~\citep{rombach2022high} for efficient spatial compression; and a Diffusion Transformer (DiT)~\citep{peebles2023scalable,wan2025}, which serves as the latent video diffusion backbone. The architectural configurations for each component are detailed in Table~\ref{tab:architecture}.

The 7B-parameter VLM, comprising a 32-layer Vision Transformer (ViT) and a 28-layer Large Language Model (LLM), is responsible for encoding textual prompts and subject images into a unified, semantically rich space. The VAE, with a 54M-parameter encoder and a 73M-parameter decoder, utilizes an $8 \times$ spatial downsampling factor, compressing video frames into a compact latent representation that is computationally tractable.

\paragraph{Subject Image Encoder}
We employ the pre-trained VAE from Wan-Video~\citep{wan2025} to transform subject images from the pixel space into the latent space. This process yields a $16$-channel latent representation with an $8\times$ downsampling factor. This latent representation is subsequently processed by a $2 \times 2$ patch embedding layer, which further reduces its spatial dimensions to align with the hidden dimension of the DiT backbone, ensuring seamless integration between the visual and generative components. The VAE weights are kept frozen during training to preserve its high-fidelity reconstruction capabilities.

\paragraph{Hierarchical Identity-Preserving Attention}
\label{apx:attention}
As detailed in the main paper, our Hierarchical Identity-Preserving Attention module is central to preventing identity leakage in multi-subject generation. It comprises three sequential stages applied within each block of the DiT, designed to progressively refine and integrate identity features. Let $Z_l$ be the video latent tokens at layer $l$, and $F_{subj} = \{f_1, f_2, \dots, f_N\}$ be the token sets for the $N$ subject images.

\paragraph{Stage 1: Intra-Subject Attention.}
To refine the feature representation for each subject independently, we first apply a standard self-attention layer~\citep{vaswani2017attention} to each subject's token set $f_i$. This initial stage allows the model to capture fine-grained, identity-specific details and consolidate features for each subject before any cross-subject interaction occurs.
\begin{equation}
    f'_i = \text{SelfAttention}(f_i) \quad \forall i \in \{1, \dots, N\}
\end{equation}

\paragraph{Stage 2: Gated Inter-Subject Attention.}
To model interactions between subjects while strictly mitigating identity bleed, we employ a novel gated cross-attention mechanism. For each subject $f'_i$, it queries the features of all other subjects $\{f'_j\}_{j \ne i}$, which then modulates the output of the cross-attention.
\begin{equation}
    f''_i = f'_i + \sigma(W_g f'_i) \odot \text{CrossAttention}(Q=f'_i, KV=\{f'_j\}_{j \ne i})
\end{equation}

\paragraph{Stage 3: Cross-Modal Attention.}
Finally, the refined subject tokens $F''_{subj}$ and the textual context tokens $f_{txt}$ attend to the main video latent tokens $Z_l$. This stage injects the rich semantic guidance from the VLM, now conditioned on the robust and disentangled subject identities. While the VLM features~\citep{Qwen2.5-VL} are computed only once, their influence is dynamic, as they modulate the video features at each layer of the DiT.
\begin{equation}
    Z'_{l} = Z_l + \text{CrossAttention}(Q=Z_l, KV=[F''_{subj}; f_{txt}])
\end{equation}
Table~\ref{tab:attention_ablation} provides a detailed ablation study on this design. These three stages form a single, cohesive block that is inserted within the first 20 layers of the DiT backbone. This design choice is based on the observation that identity features are most critical in the early stages of the generation process.

\subsection{Details of GRPO-based Post-training}\label{reward}

This section provides a more detailed breakdown of the reward calculation and hyperparameter settings for our online post-training phase, which leverages the Groupwise Policy Optimization (GRPO) algorithm~\citep{flowgrpo}.

\paragraph{Advantage Normalization}
To stabilize the training process and reduce variance in the policy gradient updates, we normalize the advantage term. The normalized advantage $\hat{A}_{i}$ for the $i$-th sample in a group of size $G$ is calculated from the total trajectory reward $r_i$:
\begin{equation}
    \hat{A}_{i} = \frac{r_{i} - \text{Mean}(\{r_{j}\}_{j=1}^G)}{\text{Std}(\{r_{j}\}_{j=1}^G) + 10^{-8}}
\end{equation}
where the total reward $r_i$ is computed by averaging the frame-level rewards over the video's duration to produce a single scalar value. The small epsilon ($10^{-8}$) is added for numerical stability. The constant $a$ for noise injection $\sigma_t$ was set to 1.0.

\paragraph{Reward Design}
As defined in the main paper, the objective is to maximize a composite reward function $\mathcal{R}_{\text{total}}$, defined as a weighted sum of fidelity and quality rewards:
\begin{equation}
    \mathcal{R}_{\text{total}}(\mathbf{V}) := w_{\text{fid}} \mathcal{R}_{\text{fid}}(\mathbf{V}, \mathcal{I}) + w_{\text{qual}} \mathcal{R}_{\text{qual}}(\mathbf{V})
\end{equation}
We empirically set the weights to $w_{\text{fid}}=0.6$ and $w_{\text{qual}}=0.4$ to prioritize identity preservation while ensuring high quality. The fidelity reward $\mathcal{R}_{\text{fid}}$ is a combination of face-specific and holistic subject consistency scores, while the quality reward $\mathcal{R}_{\text{qual}}$ assesses aesthetic appeal and physical plausibility.

\paragraph{Hyperparameters and Sensitivity}
The key hyperparameters used during the GRPO training are detailed in Table~\ref{tab:rl_hyperparams}. We also conducted a sensitivity analysis on critical parameters, with results presented in Table~\ref{tab:gamma_ablation}, to demonstrate the robustness of our training setup.

\input{tables/rl_hyperparams}

\begin{figure*}[!th]
    \centering
    \includegraphics[width=\linewidth]{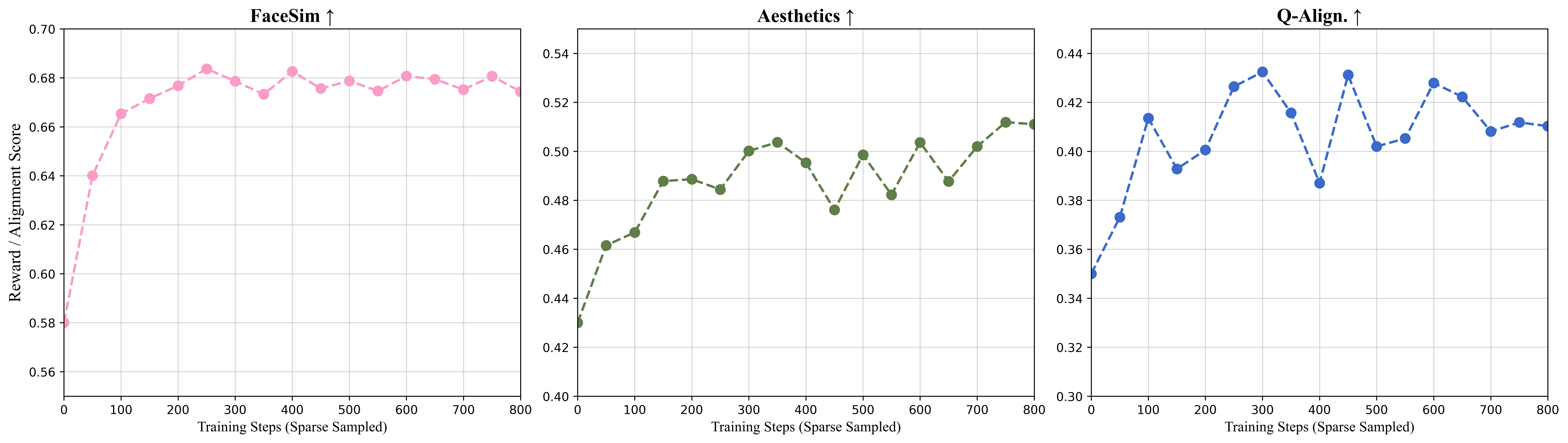}
    \caption{\textbf{Performance Curves of Online GRPO.} We plot the moving average of key evaluation metrics over 800 training steps. Our method demonstrates simultaneous improvement in identity fidelity (FaceSim), perceptual quality (Aesthetics), and text-prompt alignment (e.g., Q-Align~\cite{wu2023qalign}), validating the effectiveness of our multi-faceted reward strategy.}
    \label{fig:rl_plot}
\end{figure*}

\paragraph{Training Dynamics and Reward Effectiveness}
The effectiveness of our multi-faceted reward model is visualized in Figure~\ref{fig:rl_plot}, which illustrates the evolution of key metrics during the online GRPO phase. The plot clearly shows a positive trend for identity fidelity (e.g., FaceSim), perceptual quality (Aesthetics), and Q-Align~\cite{wu2023qalign} over the course of training. This concurrent improvement across diverse axes validates that our reward model successfully guides the generator towards producing videos that are not only more identity-consistent but also more visually appealing and semantically accurate, without suffering from catastrophic forgetting or reward over-optimization on a single metric.

\paragraph{Ablation on Fidelity Reward}
A crucial component of our total reward is the fidelity term $\mathcal{R}_{\text{fid}}$, which contains a balance factor $\gamma$ for the face-specific reward $\mathcal{R}_{\text{face}}$. Table~\ref{tab:gamma_ablation} presents an ablation study on the choice of this balance factor. The results indicate that a value of $0.5$ provides an optimal balance between average and worst-case identity fidelity, effectively preventing scenarios where one subject's identity collapses while others are preserved.

\input{tables/gamma_ablation}

\section{Dataset Curation Details}\label{apx:dataset}

\subsection{Dataset Composition and Statistics} \label{apx:dataset_stats}
% Our dataset is meticulously curated from three heterogeneous primary sources to ensure comprehensive coverage and high fidelity. The foundational component is the large-scale OpenS2V-Nexus dataset~\citep{yuan2025opens2v}, which offers a wide array of authentic scenes and actions and contributes a substantial volume of 218,230 videos, accompanied by 535,259 masked subject images and 3,098 generated counterparts. To further elevate the dataset's quality, we incorporate a high-fidelity collection of professionally shot videos with detailed annotations, comprising 9,374 pristine videos and an additional 3,000 generated subject images. This synthetically generated data, where subjects rendered by cutting-edge image editing models are placed into novel contexts, systematically increases the diversity of subject-background compositions. Finally, to provide rich priors for appearance synthesis, we integrate a dedicated Subject Image dataset. This static collection contains 14,976 images, which are further categorized into 2,877 human-centric images and 3,458 isolated clothing items. The overall composition is summarized in Table~\ref{tab:dataset_stats}, and the entire data curation pipeline is illustrated in Figure~\ref{fig:data_pipe}.
% \input{tables/data_stat}

\begin{figure*}[h!]
    \centering
    \includegraphics[width=\linewidth]{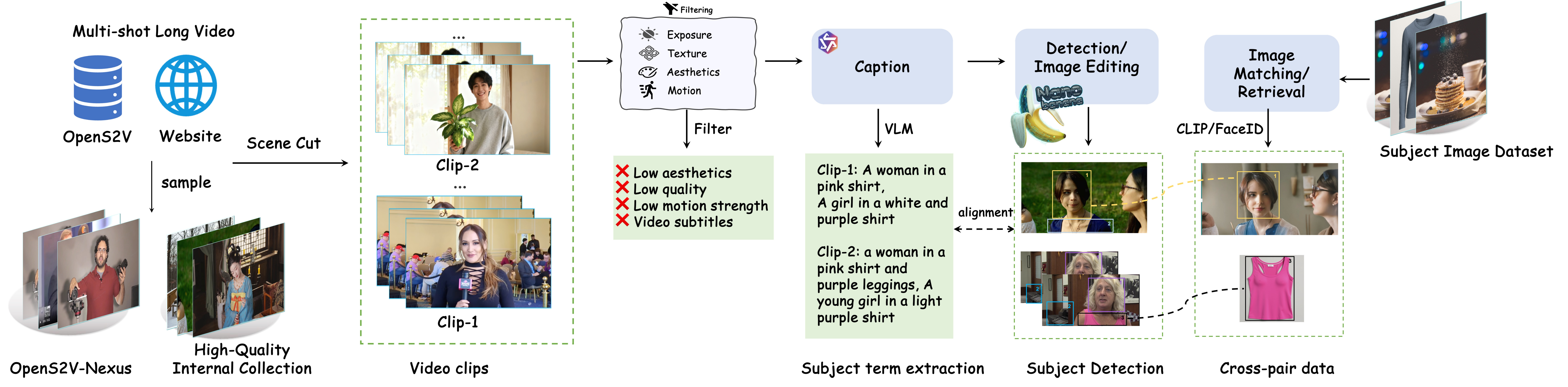}
    \caption{\textbf{Overview of the Data Curation Pipeline.} Our comprehensive data pipeline processes videos and images from diverse sources. It includes subject segmentation, high-quality reference image synthesis via an in-context learning prompt with advanced generative models, and rigorous filtering. This process yields a large-scale, high-fidelity dataset tailored for training identity-preserving video generation models.}
    \label{fig:data_pipe}
\end{figure*}

Our dataset is meticulously curated from three heterogeneous primary sources to ensure comprehensive coverage and high fidelity. The foundational component is the large-scale OpenS2V-Nexus dataset~\citep{yuan2025opens2v}, which offers a wide array of authentic scenes and actions. To ensure the temporal coherence of the clips, we incorporate a video shot detection stage~\citep{pyscenedetect}, which contributes a substantial volume of 218,230 videos, accompanied by 535,259 masked subject images and 3,098 generated counterparts. To further elevate the dataset's quality, we incorporate a high-fidelity collection of professionally shot videos with detailed annotations, comprising 9,374 pristine videos and an additional 3,000 generated subject images. This synthetically generated data, where subjects rendered by cutting-edge image editing models are placed into novel contexts, systematically increases the diversity of subject-background compositions. Finally, to provide rich priors for appearance synthesis, we integrate a dedicated Subject Image dataset. This static collection contains 14,976 images, which are further categorized into 2,877 human-centric images and 3,458 isolated clothing items. The overall composition is summarized in Table~\ref{tab:dataset_stats}, and the entire data curation pipeline is illustrated in Figure~\ref{fig:data_pipe}.
\input{tables/data_stat}

\subsection{Reference Image Construction Strategies} \label{apx:data_comp}
\begin{figure*}[!h]
    \centering
\includegraphics[width=\linewidth]{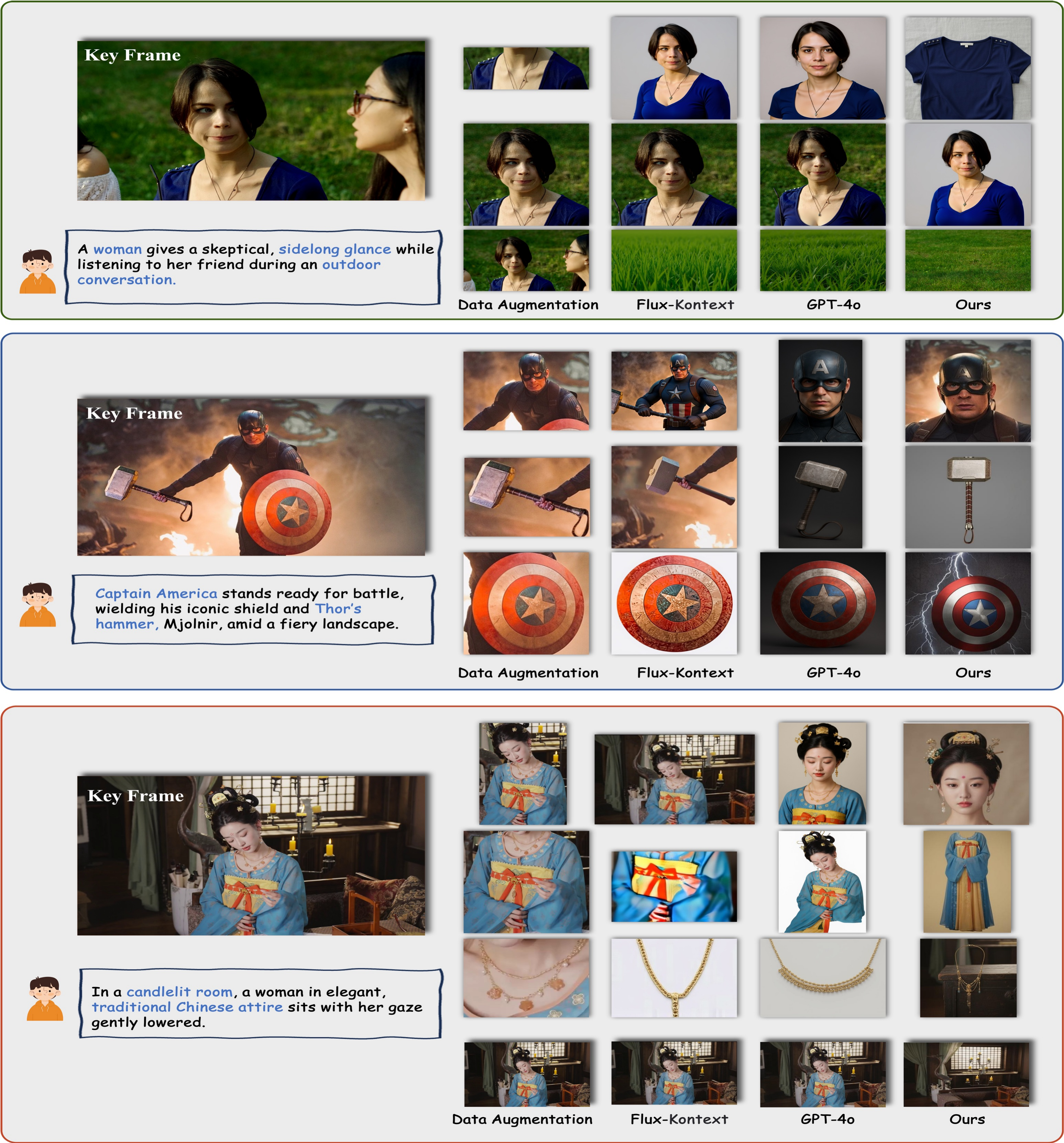}
    \caption{\textbf{Comparison of reference image construction strategies.} In contrast to direct data augmentation or image editing models like Flux-Knotext~\cite{labs2025flux} and GPT-4O~\cite{hurst2024gpt}, our data curation pipeline~\cite{Qwen2.5-VL,nanobanana} generates subject images with coherent multi-subject compositions.} 
    \label{fig:data_comparsion}
\end{figure*}

We analyze various strategies for constructing reference datasets in Figure~\ref{fig:data_comparsion}. Naive data augmentation offers limited visual diversity and is prone to generating occluded subjects, which compromises generalization to complex scenes. Alternatively, employing on-the-shelf image-consistent generators, such as Flux-Kontext~\citep{labs2025flux} or GPT-4o~\citep{hurst2024gpt}, often introduces undesirable appearance artifacts and, most critically, \textbf{fails to maintain subject fidelity}. In contrast, our approach, which leverages a tailored in-context learning prompt for the latest Gemini-Flash-Image model (a.k.a., \textcolor{blue!60}{\textbf{Nano Banana}})~\citep{nanobanana}, excels at subject decomposition and synthesis, yielding more faithful and varied reference images.

\begin{figure*}[h!]
    \centering
    \includegraphics[width=\linewidth]{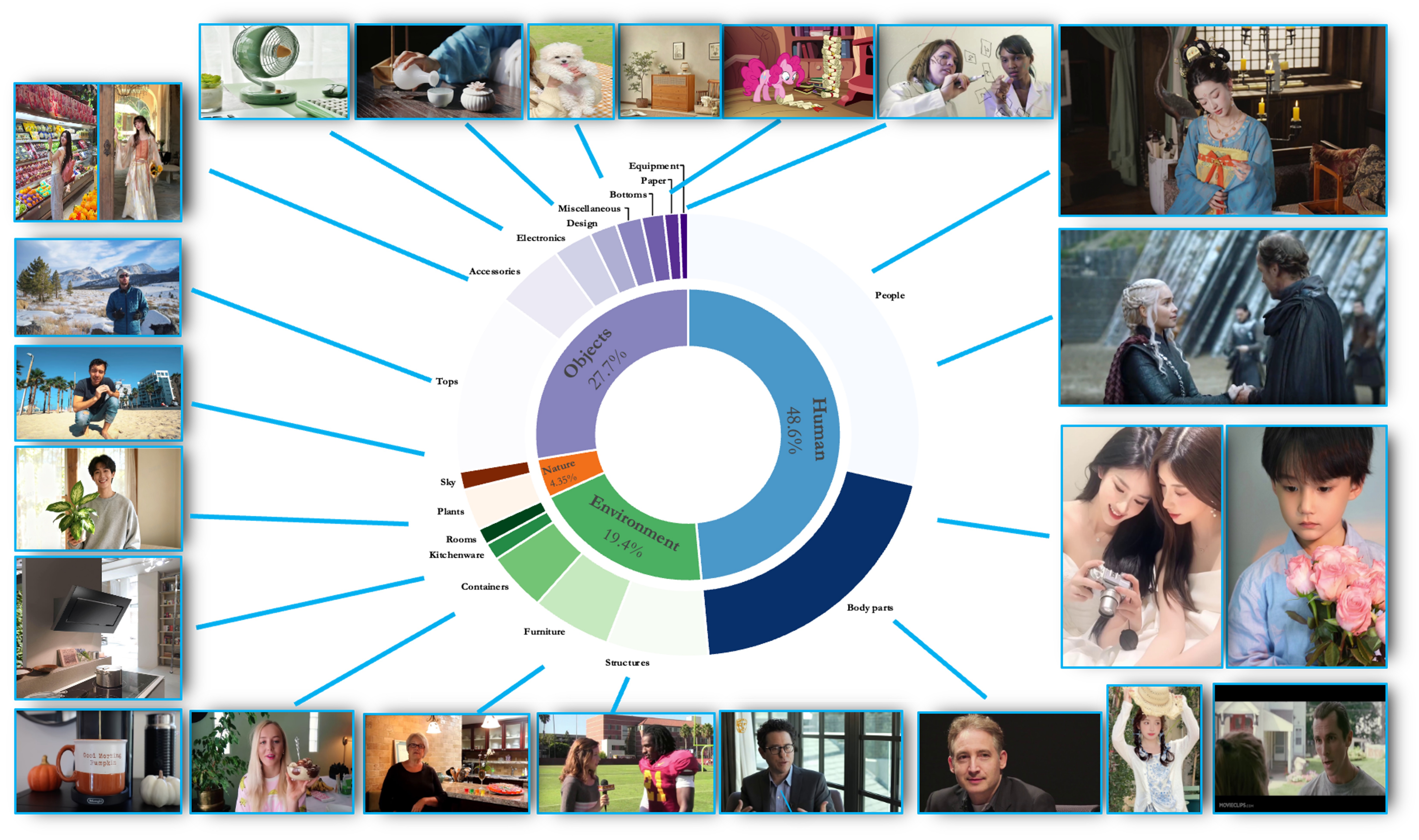}
    \caption{Statistics of the constructed dataset. The dataset is organized into four primary scenarios: Human, Objects, Environment, and Nature, each containing a variety of subcategories.}
    \label{fig:data_dist}
\end{figure*}

\begin{figure*}[h!]
    \centering
    \includegraphics[width=\linewidth]{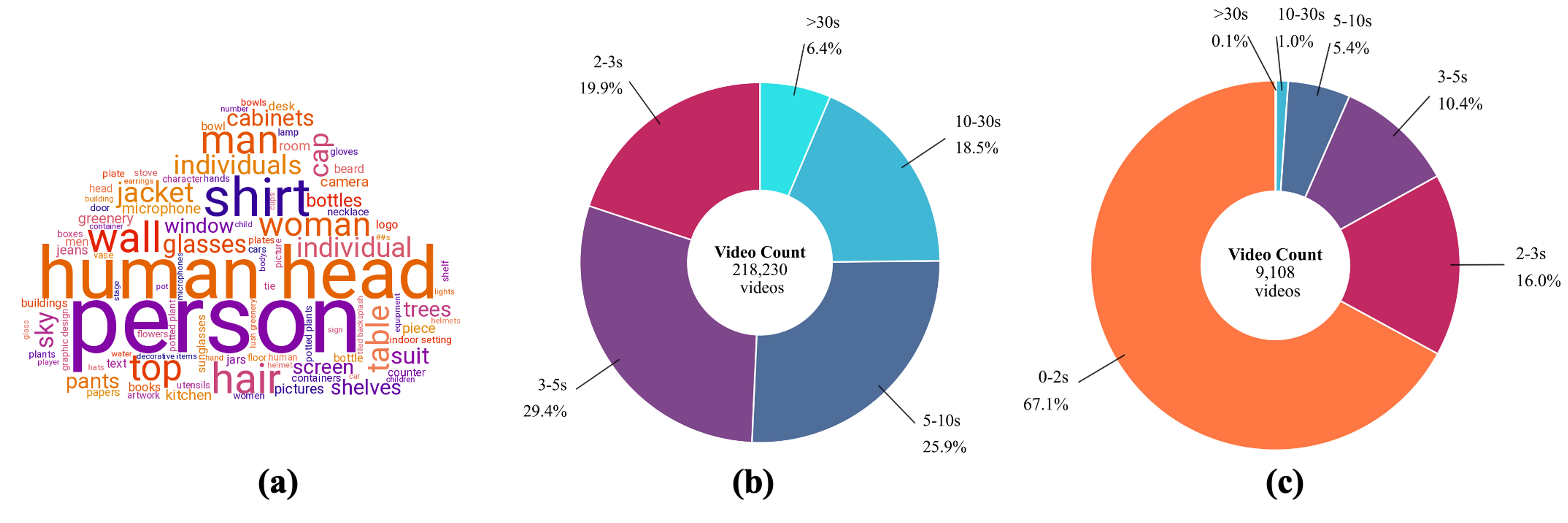}
    \caption{Statistics of the constructed dataset. (a) A word cloud illustrates the rich linguistic diversity of the reference images. The video duration distributions are shown for the two sources: (b) OpenS2V and (c) our curated videos. }
    \label{fig:data_stat}
\end{figure*}

\section{Evaluation Metric Details}\label{apx:metric}
Our evaluation protocol is principally derived from the OpenS2V-Nexus Benchmark~\citep{yuan2025opens2v}. We adopt their prescribed prompts but substitute the proprietary models with publicly available counterparts to facilitate evaluation and ensure reproducibility.

\noindent\textbf{NexusScore:} This metric quantifies the subject consistency between a generated video $\mathbf{V} = \{f_1, \dots, f_T\}$ and a reference image $\mathbf{I}_\text{ref}$. It employs a two-stage pipeline: (1) Grounded-SAM~\cite{ren2024grounded} generates a subject mask $M_t$ for each frame $f_t$. (2) A refined CLIP-based image encoder~\citep{gal2022image} $\mathcal{E}_{img}$ computes the cosine similarity between the feature embedding of the reference image and that of the masked subject in each frame. This targeted approach ensures the evaluation focuses strictly on identity fidelity by mitigating background interference. The final score is the average similarity over all $\mathbf{T}$ frames:
$ S_\text{Nexus} = \frac{1}{\mathbf{T}} \sum_{t=1}^{\mathbf{T}} \frac{\mathcal{E}_\text{img}(\mathbf{I}_\text{ref}) \cdot \mathcal{E}_\text{img}(f_t \odot M_t)}{\|\mathcal{E}_{img}(\mathbf{I}_\text{ref})\| \|\mathcal{E}_\text{img}(f_t \odot M_t)\|} $
where $\odot$ denotes element-wise multiplication.

\vspace{1em}
\noindent\textbf{NaturalScore:} This metric assesses the perceptual realism and physical plausibility of a generated video $\mathbf{V}$. We employ a state-of-the-art Multimodal Large Language Model (MLLM) to perform a deep semantic analysis. The model is presented with the video and a carefully designed prompt which instructs it to evaluate spatio-temporal consistency, identify violations of physical laws, and detect visual artifacts characteristic of generative models. The model's direct output, a normalized score, serves as the final measure of the video's naturalness.

\vspace{1em}
\noindent\textbf{GmeScore:} This metric evaluates the semantic alignment between a generated video $\mathbf{V}$ and its corresponding text prompt $\mathbf{C}$, proposed as an enhanced alternative to conventional CLIPScore. To overcome the limitations of prior methods in handling long-form text, GmeScore leverages the advanced text comprehension capabilities of the Qwen2.5-VL~\citep{Qwen2.5-VL}, which provides a powerful text encoder $\mathcal{E}_{text}$ and a temporally-aware video encoder $\mathcal{E}_{video}$. Unlike methods that average per-frame similarities, GmeScore computes a global feature representation for the entire video, thereby capturing its holistic temporal dynamics. The final score is the cosine similarity:
$ S_\text{Gme} = \frac{\mathcal{E}_\text{video}(\{f_t\}_{t=1}^T) \cdot \mathcal{E}_\text{text}(C)}{\|\mathcal{E}_\text{video}(\{f_t\}_{t=1}^T)\| \|\mathcal{E}_\text{text}(C)\|} $
This formulation provides a more accurate and comprehensive assessment of text-to-video relevance, especially for complex narratives.

% \paragraph{Reproducibility}
% To ensure reproducibility, we provide detailed experimental settings and evaluation metrics in Section~\ref{sec:implement}. Our source code, pre-trained models, curated dataset, and evaluation metrics \textcolor{blue!60}{\textbf{will be made publicly available}}.

\section{Additional Results and Analysis}\label{apx:results}

\begin{figure*}[htbp!]
    \centering
    \includegraphics[width=\linewidth]{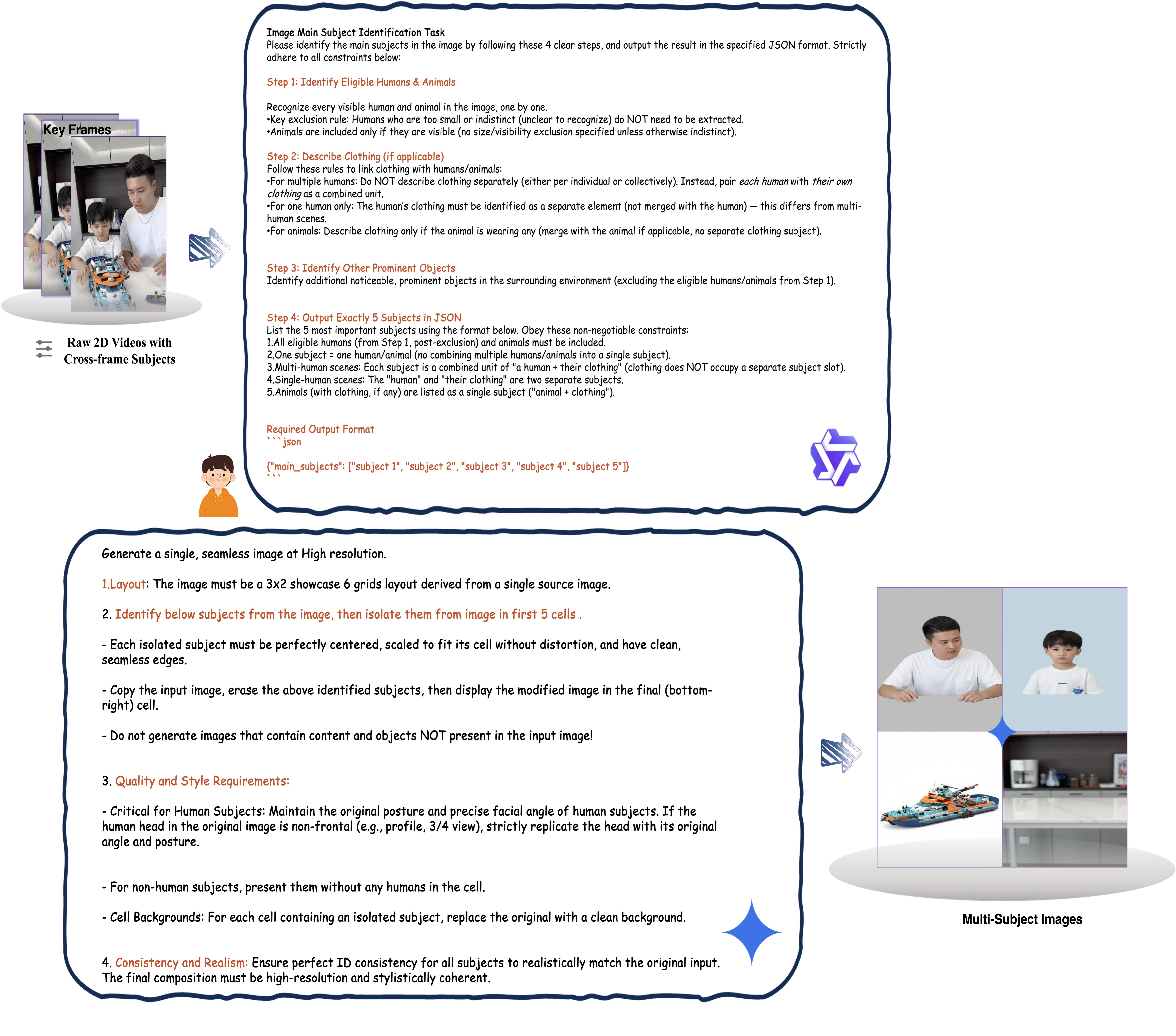}
    \caption{Prompt for MLLM-based multi-subject images generation.}
    \label{fig:prompt}
\end{figure*}

Video results for in-the-wild inputs are available in the \textcolor{blue!60}{\textbf{supplementary material Video}}.

% \subsection{Clarification on Main Results (Table 1)}
% We clarify the models presented in Table 1 of the main paper. The ``Ours-14B'' model is our large-scale model after Supervised Fine-Tuning (SFT) only, establishing a powerful SOTA baseline. The ``Ours-1.3B + online RL'' model is used to demonstrate the significant effectiveness of our proposed RLHF pipeline, as applying the full RL stage to the 14B model is computationally prohibitive.

% To demonstrate the scalability of our RL approach, we conducted an additional experiment, ``Ours-14B + RL (Light)'', where we applied RL for only 5k steps, exclusively fine-tuning the hierarchical attention blocks. As shown in Table~\ref{tab:sota_clarified}, even this lightweight RL application improves the FaceSim score from 0.412 to 0.435, a relative increase of 5.6\%, confirming the value of our method on large-scale models. The drop in FaceSim from the 1.3B+RL model (0.481) is expected due to the reduced intensity of the RL training.

% \input{tables/sota_clarified}

\subsection{Evaluation on Multi-Subject Scenarios} \label{apx:multi_subject}

\subsection{Human Preference Study} \label{apx:human_study}
\begin{figure*}[h!]
    \centering
    \includegraphics[width=0.9\linewidth]{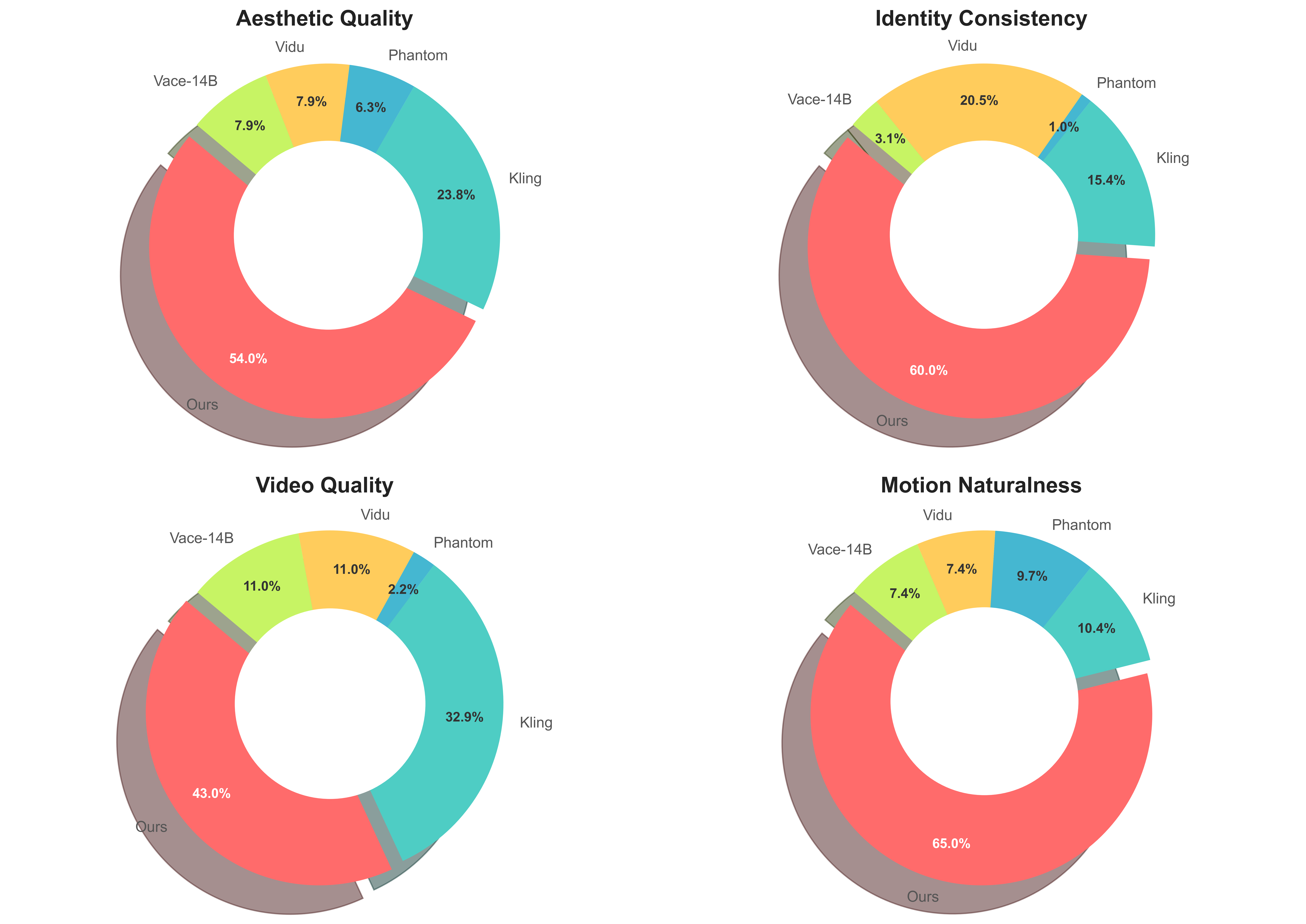}
    \caption{\textbf{Human Preference Study Results.} Participants were asked to choose the best video among five models (Ours and four competitors) based on four criteria. The chart shows the percentage of times our model was preferred. Our method significantly outperforms others in all categories, especially in Identity Consistency and Motion Naturalness.}
    \label{fig:user_study}
\end{figure*}

To address concerns of reward hacking and the potential circularity of automated metrics, we conducted an extensive human preference study. For this evaluation, a total of 200 questionnaires were gathered from 30 participants, who compared videos generated by our model against four leading competitors across four key criteria. The results, summarized in Figure~\ref{fig:user_study}, show a strong preference for our model in Aesthetic Quality (54\%), Identity Consistency (60\%), Video Quality (43\%), and Motion Naturalness (65\%), confirming its superior performance in generating high-quality, identity-consistent videos.

\section{Additional Discussion} \label{apx:discussion}

\subsection{Computational Overhead} \label{apx:computation}
The introduction of the Hierarchical Identity-Preserving Attention module adds a marginal computational overhead compared to the baseline DiT architecture. The primary increase stems from the three additional attention operations per layer for the initial 20 layers. However, since these operations are performed on the relatively small subject token sets, the impact on overall training and inference time is minimal, representing a favorable trade-off for the significant gains in identity preservation. The GRPO post-training phase requires additional computational resources, but it is a one-time cost that substantially enhances model performance.

\subsection{Ethics Statement} \label{apx:ethics}
Our work focuses on advancing controllable video generation, a technology with significant creative potential. We acknowledge the potential for misuse, such as the creation of misleading or harmful content. To mitigate these risks, we have focused our dataset curation on non-realistic, generic subjects and actions, avoiding the use of real-world public figures. The generated videos often contain subtle artifacts that distinguish them from real footage. We advocate for the development of robust detection methods for synthetic media and support the establishment of ethical guidelines for the use of generative models. Our model and dataset will be released to the research community to facilitate further research in this area, and we encourage responsible use.

%% file: tables/arch.tex
\begin{table*}[!th] 
\caption{\textbf{Detailed configuration of our model's primary architectural components}. The VLM, VAE, and DiT modules are designed to handle multimodal understanding, spatial compression, and latent diffusion, respectively.}
\centering 
\begin{tabular*}{\textwidth}{@{}l@{\extracolsep{\fill}}rrrrr@{}} 
\toprule
\textbf{Configuration} & \multicolumn{2}{c}{\textbf{VLM}} & \multicolumn{2}{c}{\textbf{VAE}} & \multicolumn{1}{c}{\textbf{DiT}} \\ 
\cmidrule(lr){2-3} \cmidrule(lr){4-5} 

& \multicolumn{1}{c}{\textbf{ViT}} & \multicolumn{1}{c}{\textbf{LLM}} & \multicolumn{1}{c}{\textbf{Encoder}} & \multicolumn{1}{c}{\textbf{Decoder}} & \\
\midrule
\# Layers & 32 & 28 & 11 & 15 & 30/40 \\
\# Num Heads (Q / KV) & 16 / 16 & 28 / 4 & - & - & 12 / 40 \\
Head Size & 80 & 128 & - & - & 128/128 \\
Intermediate Size & 3,456 & 18,944 & - & - & 8,960/13,824 \\
Patch / Scale Factor & 14 & - & 8$\times$8 & 8$\times$8 & - \\
Channel Size & - & - & 16 & 16 & - \\
\midrule
\rowcolor{blue!10}
\textbf{\# Parameters} & \multicolumn{2}{c}{\textbf{7B}} & \textbf{54M} & \textbf{73M} & \textbf{1.3B} / \textbf{14B} \\ 
\bottomrule
\end{tabular*}
\label{tab:architecture} 
\end{table*}

%% file: tables/attention_ablation.tex
% \begin{table*}[h]
% \centering
% \caption{\textbf{Ablation of Hierarchical Attention.} We analyze the contribution of each stage in our hierarchical attention mechanism. The full three-stage model provides the best performance, with \textbf{ Stage 3} (cross-modal attention to VLM) being the most critical component for semantic alignment.}
% \label{tab:attention_ablation}
% \resizebox{\textwidth}{!}{
% \begin{tabular}{lccc}
% \toprule[1.2pt]
% Method & FaceSim $\uparrow$ & NexusScore $\uparrow$ & Total Score $\uparrow$ \\
% \midrule
% \rowcolor{blue!10}
% \textbf{Full Model (3 Stages)} & \textbf{66.10\%} & \textbf{45.1\%} & \textbf{55.16\%} \\
% \midrule
% \quad w/o Stage 1 (Intra-Subject) & 63.2\% & 44.2\% & 54.1\% \\
% \quad w/o Stage 2 (Inter-Subject) & 64.5\% & 44.5\% & 54.5\% \\
% \quad w/o Stage 3 (Cross-Modal) & 58.9\% & 38.7\% & 50.8\% \\
% \quad Vanilla Attention (Stage 3 only) & 55.1\% & 39.1\% & 49.7\% \\
% \bottomrule[1.2pt]
% \end{tabular}%
% }
% \end{table*}

\begin{table*}[h]
\centering
\caption{\textbf{Ablation of Hierarchical Attention.} We analyze the contribution of each stage in our hierarchical attention mechanism. The full three-stage model provides the best performance, with \textbf{Stage 3} (cross-modal attention to VLM) being the most critical component for semantic alignment.}
\label{tab:attention_ablation}
\begin{tabular}{l ccc} % l: 左对齐, c: 居中对齐
\toprule
Method & FaceSim $\uparrow$ & NexusScore $\uparrow$ & Total Score $\uparrow$ \\
\midrule
\rowcolor{blue!10} 
\textbf{Full Model (3 Stages)} & \textbf{66.10\%} & \textbf{45.1\%} & \textbf{55.16\%} \\
\midrule
w/o Stage 1 (Intra-Subject) & 63.2\% & 44.2\% & 54.1\% \\
w/o Stage 2 (Inter-Subject) & 64.5\% & 44.5\% & 54.5\% \\
w/o Stage 3 (Cross-Modal) & 58.9\% & 38.7\% & 50.8\% \\
Vanilla Attention (Stage 3 only) & 55.1\% & 39.1\% & 49.7\% \\
\bottomrule
\end{tabular}
\end{table*}

%% file: tables/rl_hyperparams.tex
\begin{table}[h]
\centering
\caption{\textbf{Reinforcement Learning Hyperparameters.} Key hyperparameters used for the online optimization stage with our GRPO algorithm.}
\label{tab:rl_hyperparams}
\begin{tabular}{lc}
\toprule[1.2pt]
Hyperparameter & Value \\
\midrule
Algorithm & Flow-GRPO~\cite{flowgrpo,Deepseek-R1,shao2024deepseekmath} \\
Policy Learning Rate & 1e-5 \\
Optimizer & AdamW \\
AdamW $\beta$s & (0.9, 0.95) \\
AdamW $\epsilon$ & 1e-8 \\
Batch Size & 32  \\
Max Gradient Norm & 1.0 \\
\bottomrule[1.2pt]
\end{tabular}
\end{table}

%% file: tables/gamma_ablation.tex
\begin{table}[htbp] 
\centering
\caption{\textbf{Ablation on Fidelity Reward $\gamma$.} We study the effect of $\gamma$ in $\mathcal{R}_{fid}$, which balances average and minimum identity similarity. A value of 0.5 provides the best trade-off.}
\label{tab:gamma_ablation}
\resizebox{\columnwidth}{!}{
\begin{tabular}{cccc}
\toprule[1.2pt]
$\gamma$ & Avg FaceSim $\uparrow$ & Min FaceSim $\uparrow$ & Total Score $\uparrow$ \\
\midrule
0.0 (Average only) & \textbf{67.2\%} & 45.1\% & 54.8\% \\
0.25 & 66.8\% & 50.3\% & 55.0\% \\
 \textbf{0.5} & 66.1\% & \textbf{53.5\%} & \textbf{55.2\%} \\
0.75 & 65.2\% & 52.1\% & 54.9\% \\
1.0 (Minimum only) & 63.9\% & 51.5\% & 54.5\% \\
\bottomrule[1.2pt]
\end{tabular}%
} 
\end{table} 

%% file: tables/data_stat.tex
\begin{table*}[t]
\centering
\caption{\textbf{Detailed Composition of the Curated Training Dataset.} This table provides a breakdown of the three primary data sources used to train \name{}, detailing the content type, quantity, and its strategic role in the model's training.}
\label{tab:dataset_stats} % 建议重命名标签
\begin{tabularx}{\textwidth}{@{} >{\RaggedRight}X >{\RaggedRight}X r >{\RaggedRight}X @{}}
\toprule[1.2pt]
\textbf{Dataset Source} & \textbf{Content Type} & \textbf{Quantity} & \textbf{Primary Purpose} \\
\midrule

% [Row 1] 突出显示关键的预训练数据
\rowcolor{mygray}
OpenS2V-Nexus~\citep{yuan2025opens2v} & \textbf{General Videos} & \textbf{218,230} & Broad domain pre-training \\

% [Row 2] 突出显示数量最大的组件
\rowcolor{mygray}
& \textbf{Masked Subjects} & \textbf{535,259} & Learning subject-background separation \\

% [Row 3]
& Generated Subjects & 3,098 & Initial subject synthesis capability \\

\midrule

% [Row 4]
High-Quality Internal Collection & High-Fidelity Videos & 9,374 & Fine-tuning for video quality \\

% [Row 5]
& Generated Subjects & 3,000 & Enhancing appearance diversity \\

\midrule

% [Row 6] 突出显示图像数据集的总量
\rowcolor{mygray}
Subject Image Dataset & \textbf{Total Static Images} & \textbf{14,976} & Rich priors for appearance synthesis \\

% [Row 7]
& \quad - Human-centric Images & 2,877 & \\

% [Row 8]
& \quad - Isolated Clothing Items & 3,458 & \\
\bottomrule[1.2pt]
\end{tabularx}

\end{table*}